\documentclass[10pt,journal,compsoc]{IEEEtran}
\usepackage{times}
\usepackage{epsfig}

\usepackage{amssymb}

\usepackage{bm}
\usepackage{booktabs}
\usepackage{float}
\usepackage{cuted}
\usepackage{capt-of}
\usepackage{overpic}
\usepackage{multirow}
\usepackage{lipsum}
\usepackage{rotating}
\usepackage{gensymb}

\ifCLASSOPTIONcompsoc
  \usepackage[nocompress]{cite}
\else
  \usepackage{cite}
\fi
\usepackage{graphicx}

\usepackage{amsmath}
\usepackage{array}
\usepackage{url}

\usepackage[dvipsnames]{xcolor}

\begin{document}
\title{IntrinsicNGP: Intrinsic Coordinate based Hash Encoding for Human NeRF}


\author{Bo~Peng,~\IEEEmembership{Member,~IEEE,}
        Jun~Hu,~\IEEEmembership{Member,~IEEE,}
        Jingtao~Zhou,~\IEEEmembership{Member,~IEEE,}
        Xuan~Gao,~\IEEEmembership{Member,~IEEE,}
        and~Juyong~Zhang,~\IEEEmembership{Member,~IEEE,}
\IEEEcompsocitemizethanks{\IEEEcompsocthanksitem B. Peng, J. Hu, J. Zhou, X. Gao and J. Zhang are with the School
of Mathematical Science, University of Science and Technology of China.
}
\thanks{Manuscript received xx xx, 202x; revised xx xx, 202x.}}

\markboth{IEEE TRANSACTIONS ON VISUALIZATION AND COMPUTER GRAPHICS5}%
{Shell \MakeLowercase{\textit{et al.}}: Bare Demo of IEEEtran.cls for Computer Society Journals}

\IEEEtitleabstractindextext{%
\begin{abstract}
Recently, many works have been proposed to use the neural radiance field for novel view synthesis of human performers. 
However, most of these methods require hours of training, making them difficult for practical use. 
To address this challenging problem, we propose IntrinsicNGP, which can be trained from scratch and achieve high-fidelity results in a few minutes with videos of a human performer. 
To achieve this goal, we introduce a continuous and optimizable intrinsic coordinate instead of the original explicit Euclidean coordinate in the hash encoding module of InstantNGP. 
With this novel intrinsic coordinate, IntrinsicNGP can aggregate interframe information for dynamic objects using proxy geometry shapes. 
Moreover, the results trained with the given rough geometry shapes can be further refined with an optimizable offset field based on the intrinsic coordinate.
Extensive experimental results on several datasets demonstrate the effectiveness and efficiency of IntrinsicNGP. 
We also illustrate the ability of our approach to edit the shape of reconstructed objects.
\end{abstract}

}

\maketitle

\IEEEdisplaynontitleabstractindextext
\begin{figure*}[h]
    \centering
    \includegraphics[width=\linewidth]{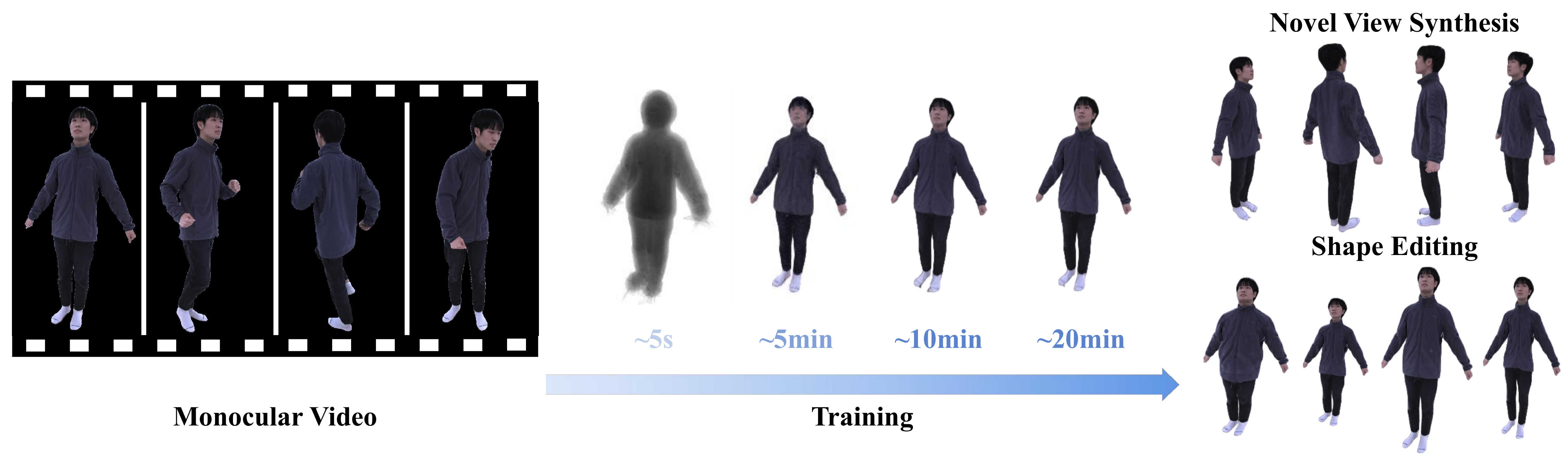}
    \vspace{-5mm}
    \captionof{figure}{For a 400-frame monocular video with a resolution of 1224 $\times$ 1024, IntrinsicNGP can train from scratch and converge within minutes on a single RTX 3090, and supports novel view synthesis as well as human shape editing.}
    \label{fig:teaser}
    \vspace{-3mm}
\end{figure*}
\IEEEpeerreviewmaketitle


\IEEEraisesectionheading{\section{Introduction}\label{sec:introduction}}
\IEEEPARstart{N}{ovel} view synthesis of human performance is an important research problem in computer vision and computer graphics. It has wide applications in many areas, such as sports broadcasting, video conferencing, and VR/AR. 
Although this problem has been widely studied for a long time, the existing methods still require quite a long computation time.
This shortcoming results in the technology not being easily adopted by public users.
Therefore, a high-fidelity novel view synthesis for human performance training within a few minutes will be of significant value for practical use.

Traditional novel view synthesis methods require dense inputs for 2D image-based methods\cite{hedman2018deep} or depth cameras for high-fidelity 3D reconstruction\cite{dou2016fusion4d} to produce realistic results.
Some model-based methods\cite{Bogo:ECCV:2016,alldieck2018video,kolotouros2019convolutional} can reconstruct explicit 3D meshes from sparse RGB videos, but they lack geometry detail and tend to be unrealistic. 
Recently, several works have applied NeRF\cite{mildenhall2020nerf} to synthesize novel views of dynamic human bodies. NeuralBody\cite{peng2021neural}, AnimatableNeRF\cite{peng2021animatable}, HumanNeRF\cite{weng_humannerf_2022_cvpr}, and other works\cite{su2021nerf,Zhao_2022_CVPR,kwon2021neural,liu2021neural, xu2021h} are able to synthesize high-quality rendering images and extract rough body geometry from sparse-view videos of the human body by combining human body priors with the NeRF model. 
However, most of these works require quite a long time to train for each subject, which is caused by the expensive computational cost of NeRF. 
Recently, INGP\cite{mueller2022instant} has improved the training speed of NeRF by several orders of magnitude with the well-designed multi-resolution hash encoding. 
However, the current strategy of INGP is based on extrinsic coordinates and works only for static scenes, and how to extend it to dynamic scenes has not yet been explored.

\begin{figure}
    \centering
    \includegraphics[width=\linewidth]{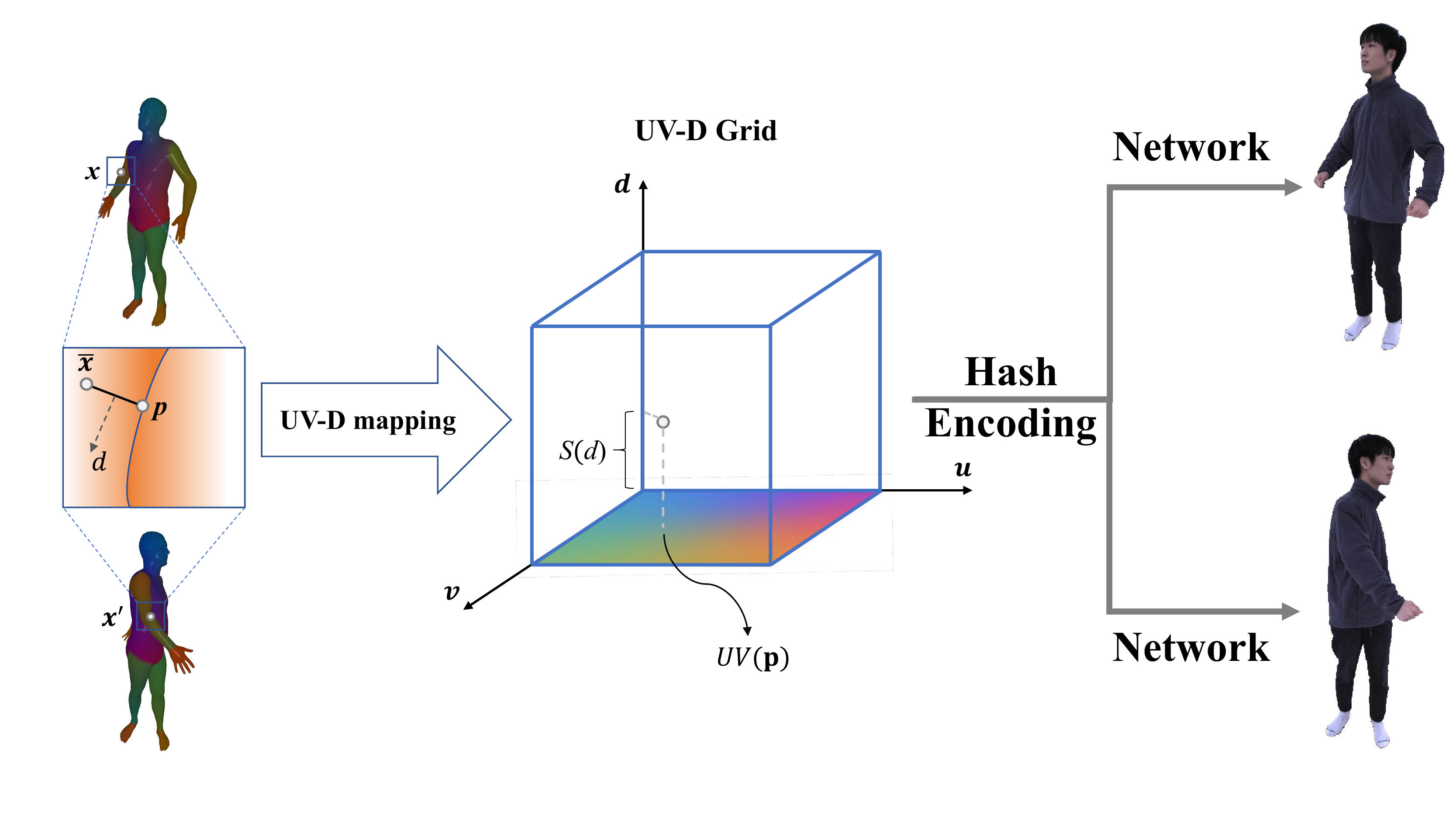}\
    \vspace{-7mm}
    \caption{\textbf{The basic idea behind IntrinsicNGP.}~ Corresponding points in different images are mapped to the same intrinsic coordinate in the UV-D grid, and thus get the same density and color via hash encoding and the NeRF network.}
    \label{fig:uvd coord}
    \vspace{-3mm}
\end{figure}

In this paper, we propose IntrinsicNGP, a novel view synthesis method for the human body, which can synthesize high-fidelity novel views of human performance even with a monocular camera, and can converge within a few minutes. 
These capabilities make IntrinsicNGP practical for common users.   
We achieve these targets through a novel intrinsic coordinate representation based on the estimated rough geometry shape, which can extend multi-resolution hash encoding to dynamic objects while aggregating information across frames.
Unlike most previous works\cite{weng_humannerf_2022_cvpr,peng2021animatable,YoungjoongKwon2021NeuralHP,RuiqiZhang2022NDFND}, which represent dynamic human bodies as a deformation field and an implicit field in canonical space, we propose an intrinsic coordinate representation independent of human motion and directly learn a radiance field based on it. The basic idea is illustrated in Fig.~\ref{fig:uvd coord}.

Specifically, given videos of human performers from sparse views or even monocular camera, we recover the rough human surface for each frame using existing model-based reconstruction methods such as EasyMocap\cite{easymocap} and VideoAvatar\cite{alldieck2018video}. For a query point in the space of the current frame, we project it onto the human surface of the current frame to obtain its nearest point on the surface. The nearest point and the corresponding signed distance value are used to represent the query point, since they remain roughly unchanged under human motion.

Although simply using the 3D coordinate of the corresponding point on the first frame's surface of the nearest point and the normalized signed distance value as the intrinsic coordinate would suffice for our needs, this representation is not easy to further optimize. 
This is because the human surface is essentially a 2D manifold and is non-convex in 3D space. Thus, it is difficult to directly optimize the 3D coordinate of a point on a surface such that the optimized point is still on the surface. 
To solve this problem, we parameterize the human surfaces as a continuous and convex 2D plane. In practice, we use the UV coordinate of the nearest point and the normalized signed distance value as our intrinsic coordinate. 
We denote the mapping from query points to intrinsic coordinates as UV-D mapping and the mapped region as UV-D grid. 
The UV-D grid records color and density information of the space around the geometry proxies, just as the UV map records texture information of its corresponding 2D manifold surface.
Moreover, since our UV-D grid is a continuous and connected domain, we optimize an offset field within this space to model detailed non-rigid deformations.
Thus, using our proposed intrinsic coordinate as the input of multi-resolution hash encoding instead of the original Euclidean coordinate, we can construct the neural radiance field for human performance based on INGP~\cite{mueller2022instant}. In addition, since IntrinsicNGP represents the human body as an explicit surface and an implicit radiance field based on it, we can render the reconstructed subject with modified body shapes by editing the input geometry proxy.

In summary, the contributions of this paper include the following aspects:
\begin{itemize}
\item We propose an intrinsic coordinate representation for hash encoding-based INGP that can aggregate interframe information, thus extending the scope of INGP from static scenes to dynamic scenes.

\item With the help of the estimated human body surface, we can reconstruct the neural radiance field of a dynamic human in a few minutes, and thus achieve high-fidelity novel view and novel shape synthesis of human performance with a monocular camera.
\end{itemize}

\section{Related Work}
Since we extend the acceleration strategy of NeRF training to dynamic objects by utilizing the human surface prior, some works related to our method are discussed in this section: human reconstruction based on NeRF, human shape reconstruction, and fast NeRF training.
\subsection{NeRF based Human Reconstruction}
NeRF(neural radiance field)\cite{mildenhall2020nerf} represents a static scene as a learnable 5D function and adopts volume rendering to render the image from any viewing direction.
Although vanilla NeRF only fits static scenes, requires dense view inputs, and is slow to train and render, much work(\cite{pumarola2021d,Niemeyer2021Regnerf,garbin2021fastnerf,park2021hypernerf,gu2021stylenerf,zhang2020nerf++,barron2021mip,tancik2022block}) has been done to solve these problems.  
Recently, a large amount of works \cite{Xu_2022_CVPR,liu2021neural,kwon2021neural,ZerongZheng2023StructuredLR,YihaoZhi2022DualSpaceNL} has focused on applying the neural radiance field to human reconstruction. 
To represent dynamic performers, NeuralBody\cite{peng2021neural} uses a set of latent codes anchored to a deformable mesh that is shared at different frames. 
And H-NeRF\cite{xu2021h} employs a structured implicit human body model to reconstruct the temporal motion of humans. 
To animate the human model, AnimatableNeRF\cite{peng2021animatable} introduces deformation fields based on neural blend weight fields to generate novel view synthesis of humans in unseen poses.
To overcome the inaccuracy of the input human poses, A-NeRF\cite{su2021nerf} jointly optimizes the human pose parameters during training.
Although these methods can achieve high-fidelity novel view synthesis results for human performers, they often require multiple views of videos and half a day of training. 
Weng et al.\cite{weng_humannerf_2022_cvpr} optimize for a NeRF representation of the human in a canonical T-pose and a motion field that maps the estimated canonical representation to each frame of the video via backward warps, so that it requires only monocular inputs from in-the-wild video.
HumanNeRF\cite{Zhao_2022_CVPR} presents a generated model that takes only an hour to refine with the input multi-view images. 
These methods typically require a long training or tuning time. In contrast, we introduce an intrinsic coordinate to extend multi-resolution hash encoding to dynamic cases, which makes our model converge in a few minutes.

\subsection{Human Shape Reconstruction}
\label{sec:shape}
Reconstructing high-quality 3D human shapes and poses from images is essential for novel view synthesis of dynamic humans.
Some traditional model-based works \cite{MANO:SIGGRAPHASIA:2017,STAR:2020,SMPL-X:2019} require only single-view RGB input. SMPLify\cite{Bogo:ECCV:2016} uses SMPL\cite{SMPL:2015} model to represent human body and obtains per-frame parameters through optimization. 
SMPL+D based method Videoavatars\cite{alldieck2018video} first computes per-frame poses using the SMPL model, then optimizes for the subject's shape in the canonical T-pose. 
Kolotouros et al.\cite{kolotouros2019convolutional} employs GraphCNN to directly regress the 3D location of the SMPL template mesh vertices, thus relaxing the heavy dependence on the model's parameter space. 
Although these methods cannot achieve high-quality results due to the limitations of the explicit parametric model, these fast-generated human surfaces can introduce human priors for implicit methods.\par
Instead of optimizing parameters per scene, some works use networks to learn human priors from ground truth data. 
PIFu\cite{saito2019pifu} concatenates the aligned feature of a pixel and the depth of a given query point as input to an MLP to obtain a 3D human occupancy field.
PIFuHD\cite{saito2020pifuhd} adds normal information to enhance the geometric details. 
StereoPIFu\cite{hong2021stereopifu} introduces the volume alignment feature and relative z-offset when giving a pair of stereo videos, which can effectively alleviate the depth ambiguity and restore the absolute scale information.
BCNet\cite{jiang2020bcnet} proposes a layered clothing representation that can support more categories of clothing and recover more accurate geometry.\par
To capture better geometry surfaces of humans, many researches\cite{alldieck2022phorhum,xiu2022icon,tiwari2021neural,rosu2022hashsdf,ShaofeiWang2023ARAHAV} represent the human body as the zero isosurfaces of a signed distance field (SDF). 
SelfRecon\cite{jiang2022selfrecon} represents the human body as a template mesh and SDF in canonical space, and uses a deformation field consisting of rigid forward LBS deformation and small non-rigid deformation to generate correspondences. 
Given monocular self-rotation RGB inputs, these methods are able to generate finer meshes of clothed humans than model-based methods. However, they still require the artist to manually adjust lighting, materials, etc. in the traditional rendering pipeline to produce realistic rendered images. Our method, on the other hand, can directly produce high-quality, novel view synthesis.

\subsection{Acceleration of Neural Radiance Field Training}
How to improve the training speed of NeRF has been widely studied(\cite{AnpeiChen2023TensoRFTR,MatthewTancik2020FourierFL,SunSC22,wang2022r2l}) since the emergence of NeRF\cite{mildenhall2020nerf}.
DS-NeRF\cite{kangle2021dsnerf} utilizes the depth information provided by 3D point clouds to speed up convergence and synthesize better results from fewer training views. 
KiloNeRF\cite{Reiser2021ICCV} adopts thousands of tiny MLPs instead of one large MLP, which could achieve real-time rendering and faster training.
Plenoxels\cite{yu_and_fridovichkeil2021plenoxels} represent a scene as a sparse 3D grid with spherical harmonics and thus can be optimized without any neural components. 
DIVeR\cite{wu2022diver} uses a similar scene representation, but applies deterministic rather than stochastic estimates during volume rendering.
Recently, INGP\cite{mueller2022instant} proposed to store the features of the voxel grid in a multi-resolution hash table and employ a spatial hash function to query the features, which can significantly reduce the number of optimizable parameters.
NeRFBlendShape\cite{Gao2022nerfblendshape} adopted a multi-level voxel field as the basis to speed up training.
These methods are only applicable to static scenes, how to apply them to dynamic objects is still a challenging problem.

\section{Method}
Given captured videos of a human performer, we want to train from scratch and converge in a few minutes, and then generate free viewpoint videos of the performer. 
First, we recover the rough human surfaces $\left\{{\mathbf{T}}_{t} \mid t=1, \ldots, N_{t}\right\}$ for each frame using EasyMocap\cite{easymocap}, where $t$ is the index of the frame, $N_{t}$ is the total number of frames.
In particular, we use the human surface of the first frame $\mathbf{T}_{1}$ as the template surface.

\begin{figure*}
    \centering
    \includegraphics[width=\textwidth]{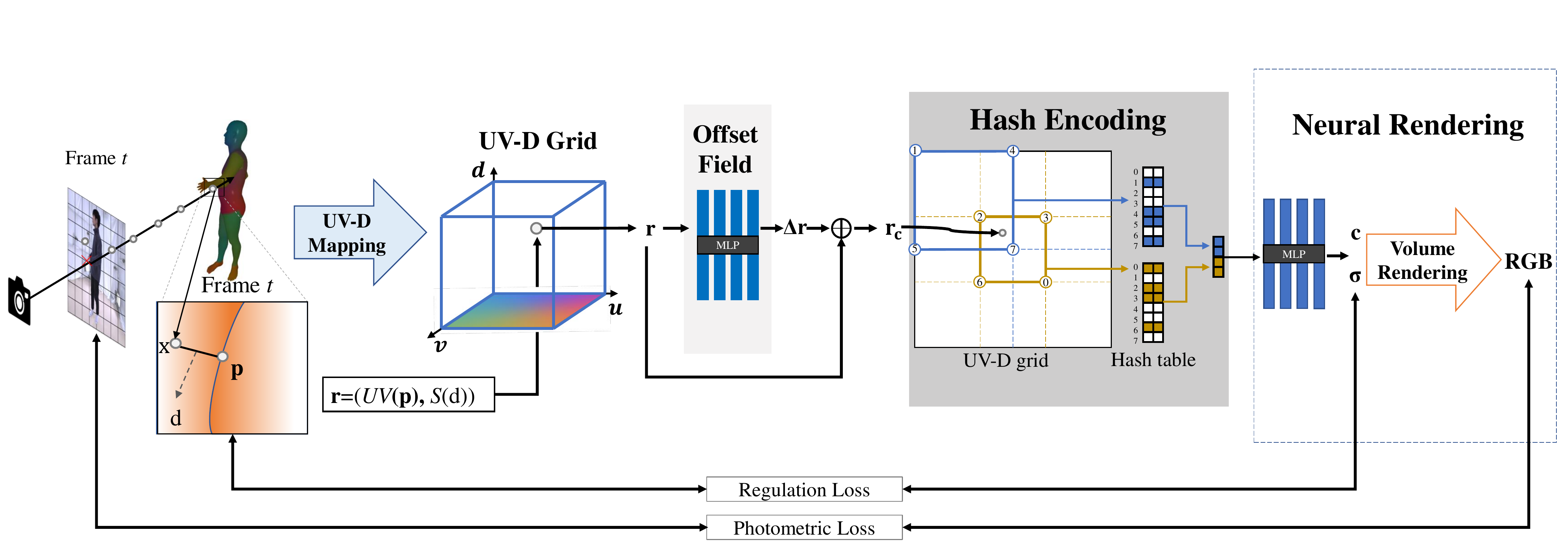}
    \vspace{-8mm}
    \caption{\textbf{Overview of IntrinsicNGP.}~ Given a sample point $\mathbf{x}$ at frame $t$, we obtain its intrinsic coordinate $\mathbf{r}$ conditioned on current human surface to aggregate the corresponding point information of different frames. An offset field is proposed to optimize the intrinsic coordinate to model detailed non-rigid deformation. Then, we use multi-resolution hash encoding to obtain the high-dimensional feature, which is the encoded input to the NeRF MLP to regress the color and density of $\mathbf{x}$.}
    \label{fig:pipeline}
    \vspace{-3mm}
\end{figure*}

Fig.~\ref{fig:pipeline} shows an overview of IntrinsicNGP. 
First, we provide some background in Sec\ref{background}. We present our intrinsic coordinate representation in Sec\ref{UVD}, which can aggregate interframe information and extend multi-resolution hash encoding to dynamic human cases. 
Hash encoding and neural radiance field based on the intrinsic coordinate are introduced in Sec\ref{representation}. Finally, our training strategy and loss are introduced in Sec\ref{loss}.

\subsection{Background}
\label{background}
\noindent{\bf{Neural Radiance Field.}} A Neural Radiance Field (NeRF)\cite{mildenhall2020nerf} represents a static 3D scene as a 5D MLP function $F_{\omega}$ that outputs the radiance emitted in each direction $\mathbf{v}$ at each point $\mathbf{x}$ in space, and a density at each point. A volume rendering strategy is then used to render images based on the neural radiance field.
In practice, for each query point $\mathbf{x}$ and view direction $\mathbf{v}$, NeRF encodes them with a positional encoding that projects a coordinate vector into a high-dimensional space. 
These high-dimensional vectors are then fed into $F_{\omega}$ to predict the density $\sigma(\mathbf{x})$ and radiance $\mathbf{c}(\mathbf{x}, \mathbf{v})$ at the input point $\mathbf{x}$ from the view direction $\mathbf{v}$. During rendering, NeRF samples a ray $\boldsymbol{\gamma} = \mathbf{o} + u\mathbf{v}$ per pixel and then calculates the color of the pixel using the following volume rendering strategy\cite{Lombardi:2019}:
\begin{equation}
\centering
\mathbf{C}(\boldsymbol{\gamma}) = \sum_{i = 1}^{N} \alpha\left(\mathbf{x}_{i}\right) \prod_{j<i}\left(1-\alpha\left(\mathbf{x}_{j}\right)\right) \mathbf{c}\left(\mathbf{x}_{i},\mathbf{v}\right),
\label{volume_rendering}
\end{equation}
where $\alpha\left(\mathbf{x}_{i}\right) = 1-\exp \left(-\sigma\left(\mathbf{x}_{i}\right) \delta_{i}\right)$, $\mathbf{x}_{i} = \mathbf{o} + u_{i}\mathbf{v}$ is the uniformly sampled point on the ray, and $\delta_{i} = {u}_{i+1}-{u}_{i}$ are the distance between adjacent sample points. 

\noindent{\bf{Multi-resolution Hash Encoding.}}
NeRF optimizes a separate neural network for each scene, and the optimization process typically takes several hours on a single RTX 3090 GPU, which is both time consuming and expensive.
To improve the training speed, INGP\cite{mueller2022instant} introduces multi-level hash encoding to replace the positional encoding in NeRF, which improves the convergence speed for a single static scene to the second level.
Specifically, INGP maintains $L$-level hash tables, and each table contains $J$ feature vectors with dimensionality $F$. We denote the feature vectors in the hash tables as 
$\mathcal{H} = \left\{{\mathcal{H}}_{l} \mid l \in\{1, \ldots, L\}\right\}$. Each table is independent, and stores feature vectors at the vertices of a grid with resolution $ N_{l}$. The resolution of each grid which is chosen according to the following strategy:
\begin{equation}
\begin{array}{c}
N_{l}:=\left\lfloor N_{\min } \cdot b^{l}\right\rfloor,\quad \quad \\
b:=\exp \left(\frac{\ln N_{\max }-\ln N_{\min }}{L-1}\right),
\end{array}
\end{equation}
where $N_{\min }$ and $N_{\max }$ are the coarsest and finest resolutions, respectively.
In practice, we set $N_{\min }=16$ and $N_{\max }=1024$.

We denote the multi-resolution hash encoding with learnable hash tables $\mathcal{H}$ as $\mathbf{h}(\cdot|\mathcal{H})$.
For a given level $l$, a 3D vector $\mathbf{z} \in [0,1]^{3}$ is scaled by the resolution of that level and then spans a voxel by rounding up and down
$ \left\lceil\mathbf{z}_{l}\right\rceil:=\left\lceil\mathbf{z} \cdot N_{l}\right\rceil$, $\left\lfloor\mathbf{z}_{l}\right\rfloor:=\left\lfloor\mathbf{z} \cdot N_{l}\right\rfloor$. 
The feature of $\mathbf{z}$ at level $l$ is tri-linearly interpolated by the feature vectors at each corner of this voxel. 
The feature vectors at each corner are queried from $\mathcal{H}_{l}$ using the following spatial hash function\cite{teschner2003optimized}: 
\begin{equation}
    hash(\mathbf{r})=\left(\bigoplus_{i=1}^{3} r_{i} \pi_{i}\right) \bmod J,
\end{equation}
where $\bigoplus$ denotes the XOR operation and $\pi_{i}$ are preset large prime numbers.
The $L$ levels' feature vectors of $\mathbf{z}$ are then concatenated to produce $\mathbf{h}(\mathbf{z}|\mathcal{H}) \in \mathbf{R}^{LF}$, which is the encoded input to the NeRF MLP.

\subsection{Intrinsic Coordinate Representation}
\label{UVD}
Given a sample point $\mathbf{x}$ at frame $t$, we expect to construct an intrinsic coordinate representation conditioned on the current surface shape $\mathbf{T}_t$. This representation aims to ensure that the corresponding points in different frames are mapped to the same intrinsic coordinate and thus obtain the same feature vectors through the multi-resolution hash encoding.
Specifically, for two points $\mathbf{x}$ and $\mathbf{x}'$ at different frames $t$ and $t'$ respectively, we should have $Map(\mathbf{x}|\mathbf{T}_t)=Map(\mathbf{x}'|\mathbf{T}_{t'})$ if they are in correspondence.
We observe that as the human body moves over time, the closest point to the query point on the human surface and the corresponding signed distance value remain approximately unchanged. Based on this observation, we use the nearest point $\mathbf{p}$ on the current human surface $\mathbf{T}_{t}$ and the corresponding signed distance $d$ to represent the query point $\mathbf{x}$ at frame $t$. 
In the following, we introduce two different intrinsic coordinate representations, XYZ-D and UV-D, based on $\mathbf{p}$ and $d$, and discuss why we choose the UV-D representation in our work.

\noindent{\bf{XYZ-D Representation.}}
\label{XYZD representation}
We first take the coordinate of the corresponding point $\bar{\mathbf{p}}$ of $\mathbf{p}$ on the template surface $\mathbf{T}_{1}$ and the signed distance value $d$ as our intrinsic coordinate, which we call the XYZ-D representation. In practice, given any sample point $\mathbf{x}$ at frame $t$, we first compute its nearest points $\mathbf{p}$ on the human surface $\mathbf{T}_t$ and the corresponding signed distance value $d$. Furthermore, we can find the corresponding point $\bar{\mathbf{p}}$ of $\mathbf{p}$ on the template surface $\mathbf{T}_{1}$ by consistent barycentric weights. Formally, we compute the XYZ-D representation as: 
\begin{equation}
\begin{array}{c}
      \rho(\mathbf{x}|\mathbf{T}_{t}) = (\mathcal{N}(Coord(\bar{\mathbf{p}})), S(d)), \\
\end{array}
\end{equation}
where $\rho(\mathbf{x}|\mathbf{T}_{t})$ is the XYZ-D representation of $\mathbf{x}$ at frame $t$, $Coord(\cdot)$ is the 3D coordinate of the input vertex. $\mathcal{N}(\cdot)$ is a linear normalization function, $S(\cdot)$ is a sigmoid function used to normalize $d$ to [0,1]. Here we use the sigmoid function to normalize $d$ since we focus on the region around $d=0$. 

\noindent{\bf{UV-D Representation.}}
\label{UVD representation}
Although the XYZ-D representation can aggregate information across frames, it is difficult to perform optimization based on it.

The XYZ of the XYZ-D representation refers to the normalized coordinate of the points on the template surface $\mathbf{T}_{1}$. 
Thus, the feasible region of this optimization problem should be $\mathcal{N}(\mathbf{T}_{1})$ × $[0,1]$, which is not convex in $[0,1]^4$.
Therefore, optimizing in $[0,1]^4$ without additional constraints will optimize outside the feasible region, leading to a meaningless result. 

To solve this problem, we map the human surface sequence to an smooth and convex region in 2D space, since the human surface is essentially a 2D manifold. 
In practice, we use the UV coordinate $UV(\mathbf{p}|{\mathbf{T}}_{t})$ of $\mathbf{p}$ and the corresponding signed distance value $d$ as the intrinsic coordinate of $\mathbf{x}$, which we denote as the UV-D representation.
Formally, we define our UV-D mapping as follows:
\begin{equation}
\mathbf{r}(\mathbf{x}|\mathbf{T}_t) = (UV(\mathbf{p}|\mathbf{T}_t),S(d)),
\end{equation}
where $\mathbf{r}(\mathbf{x}|\mathbf{T}_{t})$ is the UV-D representation of $\mathbf{x}$ at frame $t$, $UV(\cdot|{\mathbf{T}}_{t})$ refers to the UV mapping from $\mathbf{T}_t$ to a pre-calculated UV map. 
Since the human surfaces $\left\{{\mathbf{T}}_{t} \mid t=1, \ldots, N_{t}\right\}$ share the same UV map, we have $UV(\mathbf{p}|{\mathbf{T}}_{t})$=$UV(\mathbf{p}'|{\mathbf{T}}_{t'})$ if $\mathbf{p}$ and $\mathbf{p}'$ are in correspondence. Thus, our UV-D representation can aggregate interframe information as well as the XYZ-D representation. 
$\mathbf{r}(\cdot|\mathbf{T}_t)$ maps points around $\mathbf{T}_t$ to $[0,1]^3$, and we call this mapped area the UV-D grid. 
Just as the UV texture map captures the texture information of its corresponding 2D manifold, the UV-D grid captures the color and density information of the points around the 3D geometry proxies.

\begin{figure}[htb]
    \centering
    \includegraphics[width=\linewidth]{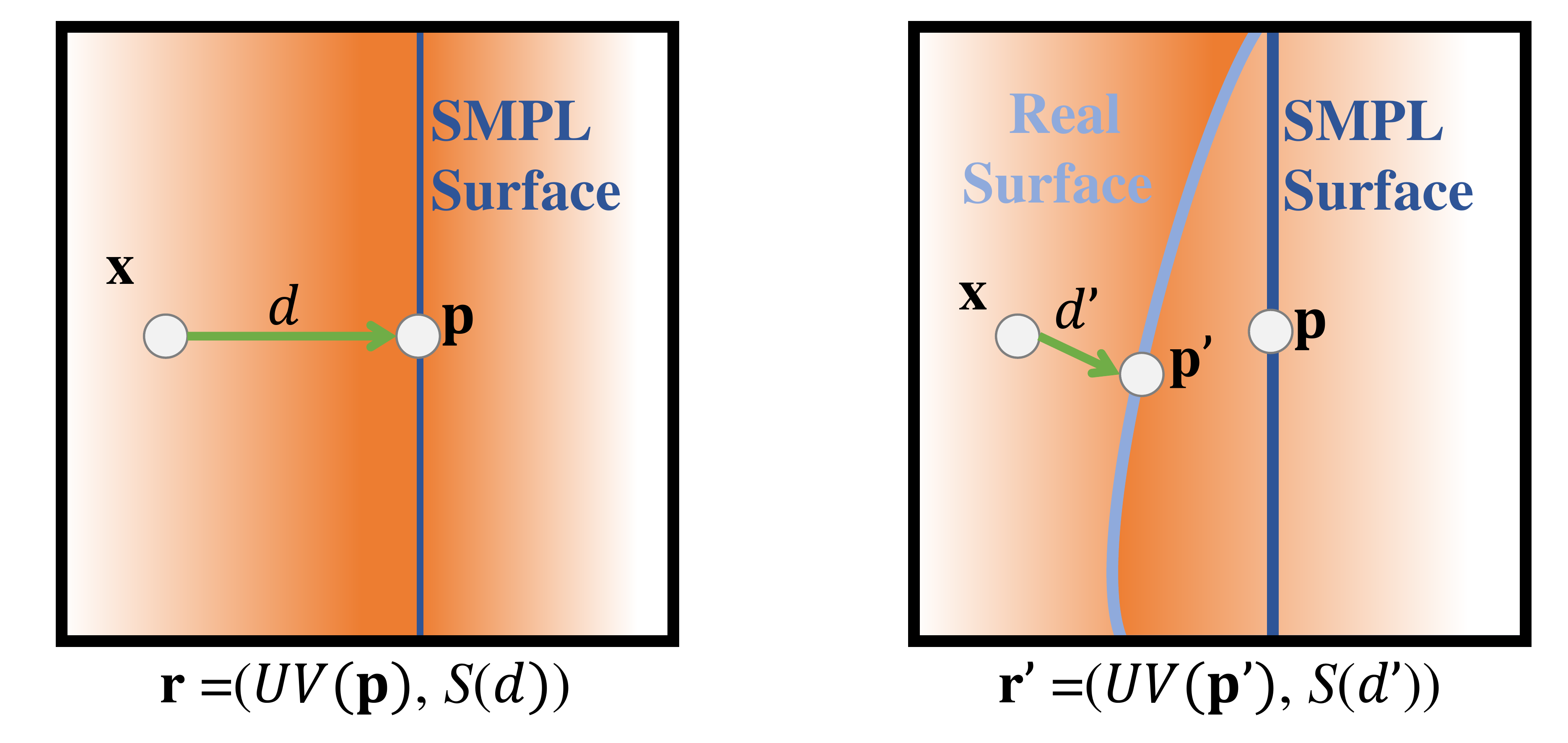}
    \vspace{-8mm}
    \caption{\textbf{The purpose of the offset field.} The rough geometry of the SMPL surface results in a biased intrinsic coordinate $\mathbf{r}$, so we use an offset field to optimize it.}
    \label{fig:wrong_mapping}
    \vspace{-5mm}
\end{figure}

\noindent{\bf{Offset Field.}}
\label{offset}
The recovered human surfaces are not always accurate enough and cannot model details, which limits the accuracy of our UV-D mapping, as shown in Fig.~\ref{fig:wrong_mapping}. To recover details, we introduce an offset field defined in the UV-D grid:
\begin{equation}
    \Delta\mathbf{r}=F_{\phi}(\mathbf{r},\mathbf{e}_{t}), 
\end{equation}
where $\mathbf{r}$ is an intrinsic coordinate, $F_{\phi}$ refers to an MLP with learnable weight $\phi$, $\mathbf{e}_{t}$ means conditional variable at frame $t$.
We also adopt the hash encoding scheme for $F_{\phi}$ to speed up training. The final optimized intrinsic coordinate of a given sample point $\mathbf{x}$ at frame t is:
\begin{equation}
    \mathbf{r_c}(\mathbf{x}|\mathbf{T}_t)=F_{\phi}(\mathbf{r}(\mathbf{x}|\mathbf{T}_t),\mathbf{e}_{t})+\mathbf{r}(\mathbf{x}|\mathbf{T}_t). 
\end{equation}

\subsection{Intrinsic NGP}
\label{representation}
Although INGP\cite{mueller2022instant} is able to converge in a short time, it only works for static scenes. To extend this fast training strategy to dynamic objects, we apply multi-resolution hash encoding based on our intrinsic coordinate instead of the original explicit Euclidean coordinate. The encoded input is then fed into the NeRF MLP $F_\omega$ to regress radiance and density. The radiance field in the UV-D grid is defined as:

\begin{equation}
    \label{eq_forward}
    \left(\sigma(\mathbf{r}), \mathbf{c}(\mathbf{r})\right) = F_{\omega}\left(\mathbf{h}(\mathbf{r}|\mathcal{H})\right),
\end{equation}
where $\sigma(\mathbf{r})$ and $\mathbf{c}(\mathbf{r})$ are the density and radiance of an intrinsic coordinate $\mathbf{r}$ in the UV-D grid, $F_{\omega}$ is the MLP function with learnable weights ${\omega}$.

The density and radiance of a query point $\mathbf{x}$ at frame $t$ can now be defined as:
\begin{equation}
    \left(\sigma_{t}(\mathbf{x}), \mathbf{c}_{t}(\mathbf{x})\right) =  F_{\omega}\left(\mathbf{h}(\mathbf{r_c}(\mathbf{x}|\mathbf{T}_t)|\mathcal{H})\right),
\end{equation}

Finally, the color of each sample ray is calculated using the following volume rendering formula:
\begin{equation}
\centering
\mathbf{C}_{t}(\boldsymbol{\gamma}) = \sum_{i = 1}^{N} \alpha_{t}\left(\mathbf{x}_{i}\right) \prod_{j<i}\left(1-\alpha_{t}\left(\mathbf{x}_{j}\right)\right) \mathbf{c}_{t}\left(\mathbf{x}_{i}\right),
\label{volume_rendering2}
\end{equation}
where $\alpha_{t}\left(\mathbf{x}_{i}\right) = 1-\exp \left(-\sigma_{t}\left(\mathbf{x}_{i}\right) \delta_{i}\right)$. 
As with HumanNeRF\cite{weng_humannerf_2022_cvpr}, we did not include the view direction $\mathbf{v}$ as an input. This is because it could lead to color bias in the novel view synthesis when inputting monocular videos.

\subsection{Training}
\label{loss}
We use the following loss function to jointly optimize the network parameters ${\omega, \phi}$ and the feature vectors $\mathcal{H}$ in the hash tables.

\subsubsection{Photometric Loss}
We minimize the rendering error of all observed images, and the loss function is defined as:
\begin{equation}
L_{\mathrm{rgb}}=\frac{1}{|\mathcal{R}|}\sum_{\boldsymbol{\gamma} \in \mathcal{R}}\left\|\tilde{\mathbf{C}_{t}}(\boldsymbol{\gamma})-\mathbf{C}_{t}(\boldsymbol{\gamma})\right\|_{2},
\end{equation}
where $\mathcal{R}$ is the set of rays passing through image pixels and $\tilde{\mathbf{C}_{t}}(\boldsymbol{\gamma})$ is the corresponding ground truth color.

\subsubsection{Regulation Loss}
\label{reg_loss}
\noindent{\bf{Mask Loss.}} We require the weight sum of the ray to match the masks, which is obtained by MiVOS\cite{cheng2021mivos} from the input videos.
\begin{equation}
L_{\mathrm{mask}}=\frac{1}{|\mathcal{R}|}\sum_{\boldsymbol{\gamma} \in \mathcal{R}}\left ({\mathbf{W}_{t}}(\boldsymbol{\gamma})(1-\mathbf{M}(\boldsymbol{\gamma}))+(1-\mathbf{W}_{t}(\boldsymbol{\gamma}))\mathbf{M}(\boldsymbol{\gamma})\right ),
\end{equation}
where $\mathbf{W}_{t}(\boldsymbol{\gamma}) = \sum_{i = 1}^{N} \alpha_{t}\left(\mathbf{x}_{i}\right) \prod_{j<i}\left(1-\alpha_{t}\left(\mathbf{x}_{j}\right)\right)$ is the weight sum of the ray $\boldsymbol{\gamma}$, $\mathbf{M}(\boldsymbol{\gamma})=1$ if $\boldsymbol{\gamma}$ is sampled from the pixel in the masked area otherwise 0.

 \begin{figure*}
    \centering
     \includegraphics[width=\linewidth]{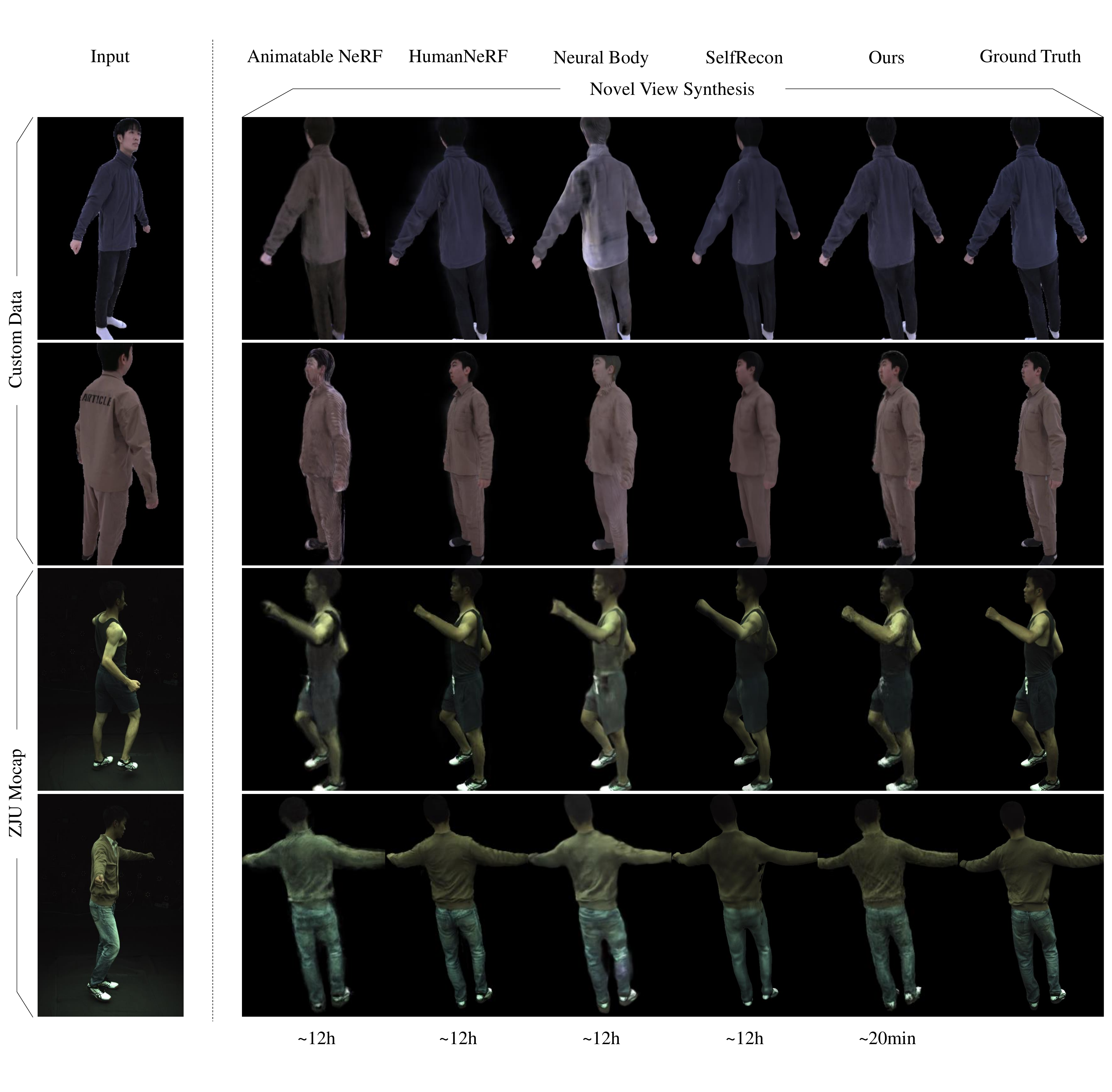}
    \vspace{-14mm}
    \caption{Qualitative comparison result on the ZJU-MoCap dataset and our custom data. Our method produces satisfactory results with much less training time. The bottom row are the training time of each method.}
    \label{fig:compare}
    \vspace{-3mm}
\end{figure*}

\noindent{\bf{Distance Loss.}} If a point is far away from the human body, its density should be close to 0. Therefore, we use an exponential function to penalize the density outside the human body.
\begin{equation}
L_{\mathrm{dist}}=\frac{1}{|\mathcal{X}|}\sum_{\mathbf{x} \in \mathcal{X}} \sigma_{t}(\mathbf{x}) \exp(\varphi(d)\beta),
\end{equation}
where $\mathcal{X}$ is the set of points sampled on the rays in $\mathcal{R}$, $\beta$ is a hyperparameter,  and $\varphi(\cdot)$ refers to the Relu function.

\noindent{\bf{Offset Regularization Loss.}}
The output of the offset field should be relatively small. So we use the following:
\begin{equation}
L_{\mathrm{dfm}}=\frac{1}{|\mathcal{X}|}\sum_{\mathbf{x} \in \mathcal{X}} \left\| F_{\phi}(\mathbf{r}(\mathbf{x}|\mathbf{T}_{t}),\mathbf{e}_{t}) \right\|_{2}.
\end{equation}

\noindent{The final regulation guidance loss is defined as:}
\begin{equation}
    L_{\mathrm{reg}}=\lambda_{\mathrm{mask}}L_{\mathrm{mask}}+\lambda_{\mathrm{dist}} L_{\mathrm{dist}}+\lambda_{\mathrm{dfm}}L_{\mathrm{dfm}},
\end{equation}

 \begin{figure*}
    \centering

     \includegraphics[width=\linewidth]{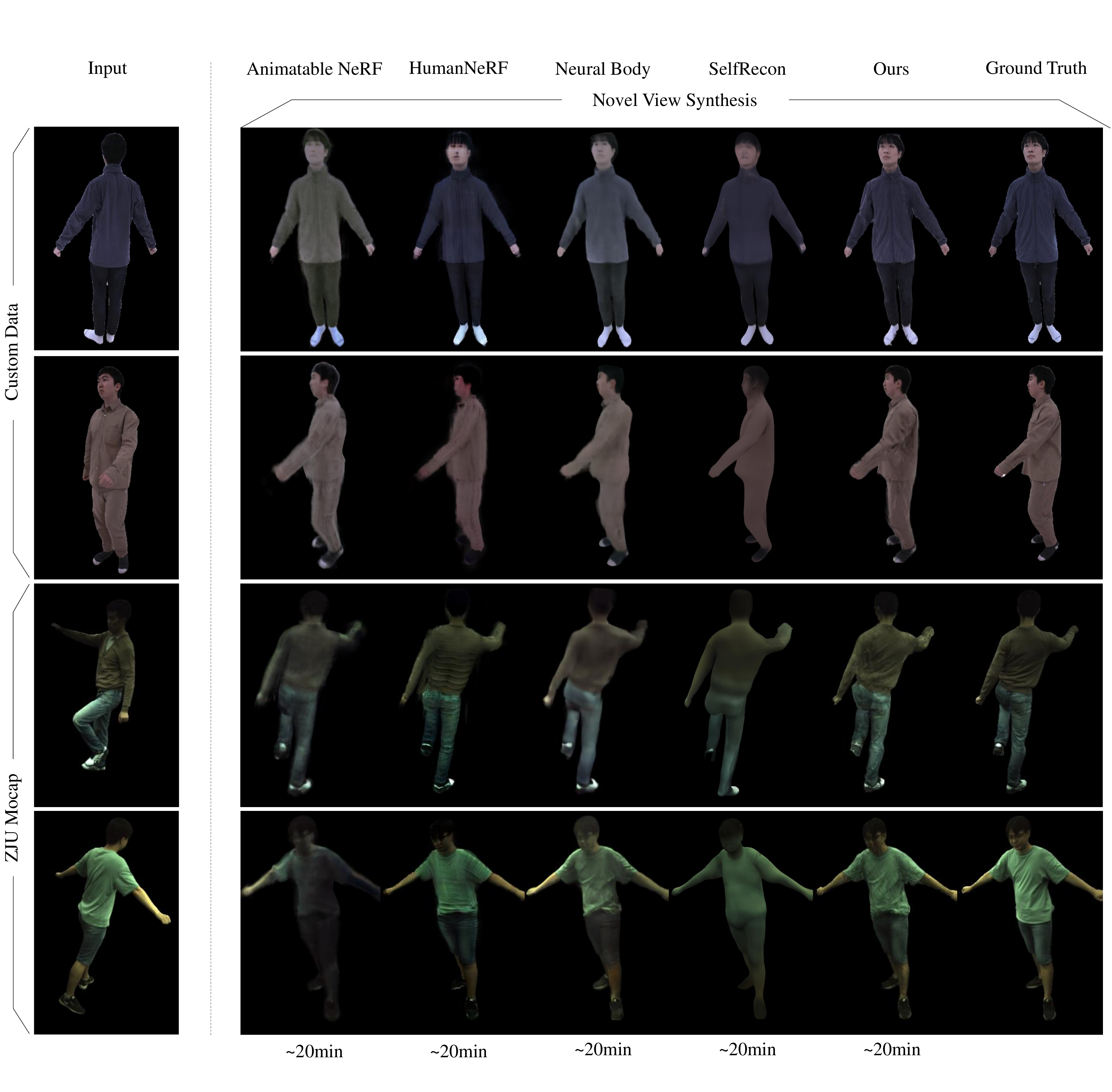}

    \vspace{-8mm}
    \caption{Qualitative comparison result for monocular case with 20 minutes of training. Our method can produce high fidelity results within 20 minutes, while other methods produce results with incorrect geometry or textures. The bottom row shows the training time for each method.}
    \label{fig:compare_20min}
    \vspace{-3mm}
\end{figure*}

\subsubsection{Training Strategy}
The total loss function is formulated as:
\begin{equation}
     L_{\mathrm{total}} = L_{\mathrm{rgb}} + \lambda L_{\mathrm{reg}}.
\end{equation}
We set $\lambda=1$ for the first 400 iterations to learn a coarse geometry of the human body, and then set $\lambda = 0.1$ to learn the details mainly through photometric loss.



\subsection{Implementation Details}
\label{detail}
We implement our code on top of the torch-ngp\footnote{https://github.com/ashawkey/torch-ngp} codebase\cite{tiny-cuda-nn,mueller2022instant,torch-ngp,tang2022compressible}.
We optimize with Adam\cite{DBLP:journals/corr/KingmaB14} using a learning rate decay from $2 \cdot 10^{-3}$ to $2 \cdot 10^{-5}$. 
To speed up the evaluation, we render a wide mask of the human performer with the recovered human surfaces and apply volume rendering only in this region. 
The convergence time of our method depends on the length and resolution of the input videos.
For a 1080$\times$1080 monocular video with 200 frames, we need about 3K iterations to converge (about 12 minutes on a single NVIDIA GeForce RTX3090).
In addition, because we use multi-resolution hash encoding to speed up training,
our model requires additional memory to store the hash tables. 
Our model is about twice the size of an INGP model because we use two hash encoders in our pipeline.

\begin{figure*}
    \centering
    \label{compare_multi}

     \includegraphics[width=\linewidth]{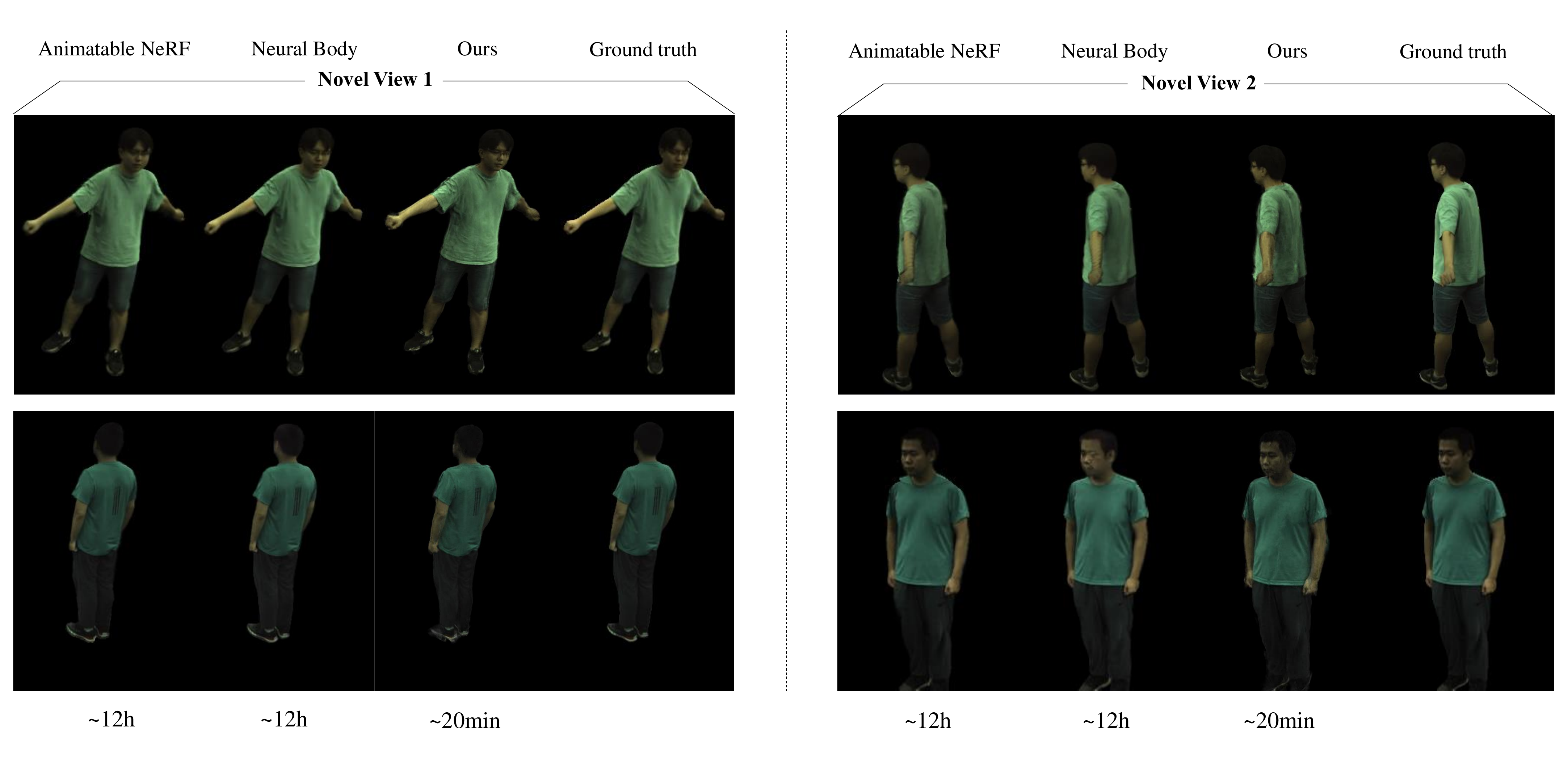}

    \vspace*{-10mm}
    \caption{Qualitative comparison result on the ZJU-MoCap dataset for multi-view inputs. Our method produces better novel view synthesis results with much less training time. The bottom row shows the training time of each method.}
    \label{fig:comparemulti}
  
\end{figure*}

\section{Experiments}
To demonstrate the effectiveness of our method, we perform comparative experiments on both single-view and multi-view videos. Some ablation studies are also discussed to evaluate the necessity of our modules. And we also show an application of our method.
\subsection{Dataset}
\label{data}
To evaluate the reconstruction capability of our method from single-view input, we capture some custom data including monocular videos of human performers and the corresponding SMPL meshes. 
Each video contains the whole-body information of the performer from a single view.
Videos from other views are provided for evaluation. 
We also use the ZJU-Mocap\cite{peng2021neural} dataset for comparison with state-of-the-art methods.

\noindent{\bf{metrics.}} Following NeRF\cite{mildenhall2020nerf}, we use two standard metrics to quantify the results of novel view synthesis: peak signal-to-noise ratio (PSNR) and structural similarity index (SSIM). 
To reduce the effect of background, we compute these metrics only for the pixels inside the 2D bounding box obtained from the input mask for each view.

\subsection{Comparison}

\subsubsection{Results on Monocular Videos}
We compare with state-of-the-art methods for implicitly human novel view synthesis from monocular inputs:

1) AnimatableNeRF\cite{peng2021animatable} uses deformation field based on neural blend weight to aggregate per-frame information to reconstruct the neural radiance field in canonical space.

2) NeuralBody\cite{peng2021neural} reconstructs per-frame neural radiance field conditioned on body-structured latent codes, which are diffused into the whole space via SparseConvNet.

3) SelfRecon\cite{jiang2022selfrecon} reconstructs the clothed human body by combining implicit and explicit representations to recover coherent space-time geometries. 

4) HumanNeRF\cite{weng_humannerf_2022_cvpr} decomposes human motion into skeletal motion and pose-based non-rigid motion, and then reconstructs the neural radiance field of canonical space.

We perform this experiment on ZJU-Mocap\cite{peng2021neural} and our custom data. 
Specifically, we select 5 subjects (313, 377, 386, 392, 394) from ZJU-Mocap and 3 subjects from our custom data with relatively high image quality and use ``Camera (1)'' for training and other views for evaluation.
We use the official open source code of these methods for comparison with our method. 

As shown in Fig.~\ref{fig:compare}, SelfRecon, HumanNeRF, and our method can produce satisfactory results similar to the ground truth even on completely unobserved views, while NeuralBody and AnimatableNeRF tend to produce results with blur and color differences. 
Tab.~\ref{tab:compare_mono} shows the quantitative results for training frames and novel poses, respectively. 
Our method outperforms NeuralBody and AnimatableNeRF for all subjects and all metrics. 
HumanNeRF and SelfRecon outperform us by a small margin on PSNR and SSIM, respectively, partly because we rely entirely on SMPL to model skeletal motion, while HumanNeRF optimizes the blend weights. 
And SelfRecon is based on surface rendering, which is less likely to generate noise compared to volume rendering. 
We believe that IntrinsicNGP can perform better with finer human surfaces. 
It should be noted that our method takes only about twenty minutes to train each model on a single RTX3090, while the other methods take twelve hours.

\begin{table}[t]
    \caption{Quantitative comparison for the monocular case.}
    \vspace{-6mm}
    \label{tab:compare_mono}
    \begin{center}
    \resizebox{\linewidth}{!}{
    \begin{tabular}{|ccccc|}
    \hline
    Method      & Dataset  & PSNR $\uparrow$ & SSIM $\uparrow$   & Training time $\downarrow$  \\
    \hline
    \multirow{2}*{NeuralBody~\cite{peng2021neural}}&ZJU-Mocap &24.15 
    &0.865 &\multirow{2}*{$\sim$12 h}\\
    &Custom data &22.98 &0.853 &\\
    \hline
    \multirow{2}*{AnimatableNeRF~\cite{peng2021animatable}}&ZJU-Mocap &24.30 &0.854 &\multirow{2}*{$\sim$ 12h}\\
    &Custom data &23.14 &0.829 &\\
    \hline
    \multirow{2}*{HumanNeRF~\cite{weng_humannerf_2022_cvpr}}&ZJU-Mocap &\textbf{25.99} &0.879 &\multirow{2}*{$\sim$ 12h}\\
    &Custom data &\textbf{26.38} &0.836 &\\
    \hline
    \multirow{2}*{SelfRecon~\cite{jiang2022selfrecon}}&ZJU-Mocap &25.20 &\textbf{0.896} &\multirow{2}*{$\sim$12h}\\
    &Custom data &25.16 &\textbf{0.905} &\\
    \hline
     \multirow{2}*{Ours}&ZJU-Mocap &25.86 &0.889 &\multirow{2}*{\textbf{$\sim$20min}}\\
     &Custom data &26.21 &0.881 &\\
    \hline
    \end{tabular}}
    \end{center}

    \vspace{-8mm}
 
\end{table}

We provide another comparison experiment to demonstrate our convergence speed. In this experiment, all baseline methods are trained for only 20 minutes.
Fig.~\ref{fig:compare_20min} shows the results of the comparison.
It can be seen that our method is able to produce high fidelity results, while AnimatableNeRF and SelfRecon fail to produce clear results. In addition, the results of NeuralBody and HumanNeRF lack details of the clothing. 
As shown in Tab.~\ref{tab:compare_20min}, our method outperforms all baseline methods when trained for only 20 minutes.

\begin{table}[t]
    \caption{Quantitative comparison for the monocular case with 20 minutes of training.}
    \vspace{-6mm}
    \label{tab:compare_20min}
    \begin{center}
    \resizebox{\linewidth}{!}{
    \begin{tabular}{|ccccc|}
    \hline
    Method      & Dataset  & PSNR $\uparrow$ & SSIM $\uparrow$   & Training time $\downarrow$  \\
    \hline
    \multirow{2}*{NeuralBody~\cite{peng2021neural}}&ZJU-Mocap &23.64 
    &0.844 &\multirow{2}*{\textbf{$\sim$ 20min}}\\
    &Custom data &22.47 &0.837 &\\
    \hline
    \multirow{2}*{AnimatableNeRF~\cite{peng2021animatable}}&ZJU-Mocap &21.68 &0.841 &\multirow{2}*{\textbf{$\sim$ 20min}}\\
    &Custom data &22.69 &0.816 &\\
    \hline
    \multirow{2}*{HumanNeRF~\cite{weng_humannerf_2022_cvpr}}&ZJU-Mocap &24.56 &0.860 &\multirow{2}*{\textbf{$\sim$ 20min}}\\
    &Custom data &23.88 &0.820 &\\
    \hline
    \multirow{2}*{SelfRecon~\cite{jiang2022selfrecon}}&ZJU-Mocap &23.36 &0.865 &\multirow{2}*{\textbf{$\sim$ 20min}}\\
    &Custom data &23.25 &0.883 &\\
    \hline
     \multirow{2}*{Ours}&ZJU-Mocap &\textbf{25.86} &\textbf{0.889} &\multirow{2}*{\textbf{$\sim$ 20min}}\\
     &Custom data &\textbf{26.21} &\textbf{0.881} &\\
    \hline
    \end{tabular}}
    \end{center}
    \vspace{-6mm}
\end{table}

\subsubsection{Results on Multi-view Videos}
\begin{table}[h]
    \caption{Quantitative comparison for multi-view inputs.}
    \vspace{-6mm}
    \begin{center}
    \resizebox{\linewidth}{!}{
    \begin{tabular}{|cccc|}
        \hline
        Method & PSNR $\uparrow$& SSIM$\uparrow$ &Training time$\downarrow$\\
        \hline
        Neural Body\cite{peng2021neural}  & 28.36 &0.901&$\sim$12h \\
        \hline
        AnimatableNeRF\cite{peng2021animatable} &  28.14 &0.895&$\sim$12h\\
        \hline
        Ours& \textbf{28.71} &\textbf{0.912}&\textbf{$\sim$ 20min}\\
        \hline
    \end{tabular}}
    \end{center}
    \label{tab:compare_multi}
    \vspace{-4mm}
\end{table}

To demonstrate that our approach can achieve higher quality results with multi-view input, we perform multi-view experiments on the ZJU-Mocap dataset.
We compare it to AnimatableNeRF\cite{peng2021neural} and NeuralBody\cite{peng2021neural}, which are state-of-the-art methods for implicitly human novel view synthesis from sparse multi-view inputs.
We conducted this experiment on the ZJU-Mocap dataset including subjects 313, 377, 386, 392, and 394.
We use the same subjects as in the monocular case and select six equally spaced views for training.

As shown in Fig.~\ref{fig:comparemulti}, our method can produce high-quality results as well as NeuralBody\cite{peng2021neural} and AnimatableNeRF\cite{peng2021neural}, but with much less training time. The results of our method have reasonable geometry details and textures.
Tab.~\ref{tab:compare_multi} shows the quantitative results.
Our method outperforms \cite{peng2021animatable,peng2021neural} on all metrics with much less training time.

\subsection{Ablation Study}

\subsubsection{Ablation Study on the Multi-resolution Hash Encoding}
\begin{figure}
    \centering
    \includegraphics[width=\linewidth]{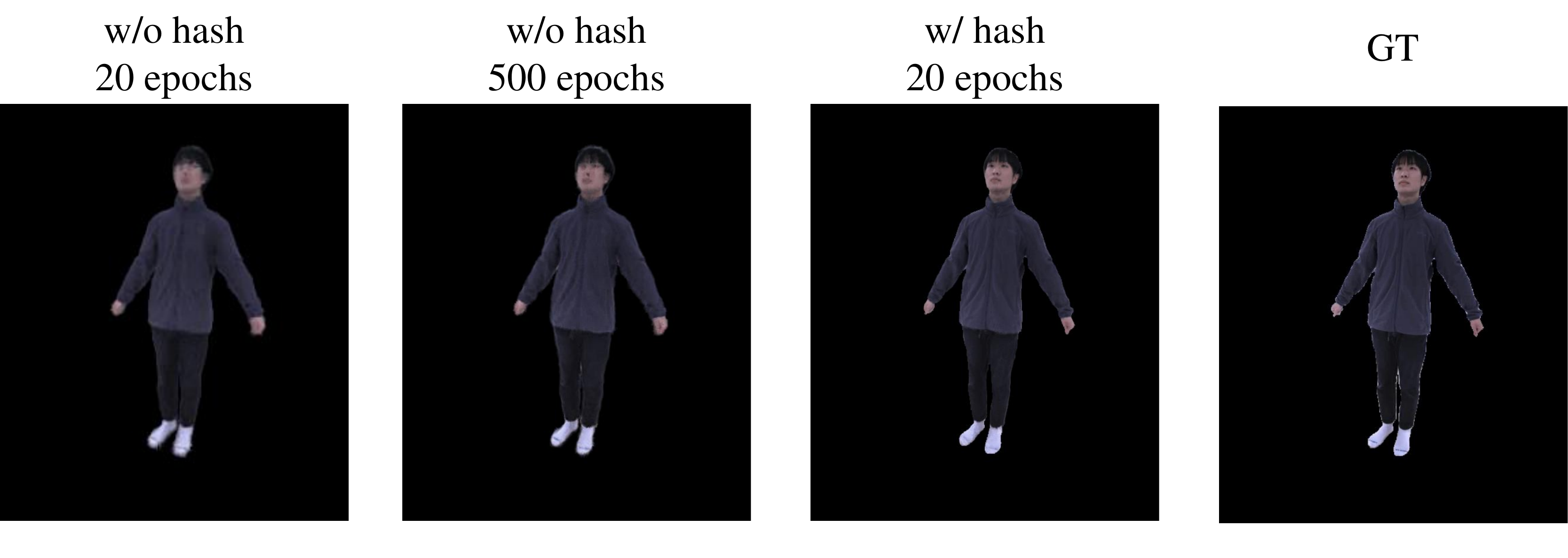}
    \put(-105, -3){\footnotesize  $\sim$20min}
    \put(-167.5, -3){\footnotesize  $\sim$6h}
    \put(-240, -3){\footnotesize  $\sim$15min}
    \vspace{-3mm}
    \caption{Ablation study on multi-resolution hash encoding. With hash encoding, our method requires much fewer training epochs to converge. The bottom row shows the training times.}
    \label{fig:ablation_hash}
\end{figure}

To further verify the effectiveness of hash encoding in our method, we design the following baseline version. 
All feature vectors originally obtained by hash encoding are now independent and optimizable.
In practice, we subdivide the UV-D grid into a $512\times512\times512$ voxel grid and anchor the feature vectors with dimension 32 directly to each grid point. We then use our custom dataset to train the original model and this pipeline. 
The corresponding rendering results are shown in Fig.~\ref{fig:ablation_hash}.
This representation takes at least twenty times longer to converge. Using multi-resolution hash encoding dramatically increases the training speed of our method.

\subsubsection{Ablation Study on the Offset Field}
\begin{figure}
    \centering
    \includegraphics[width=\linewidth]{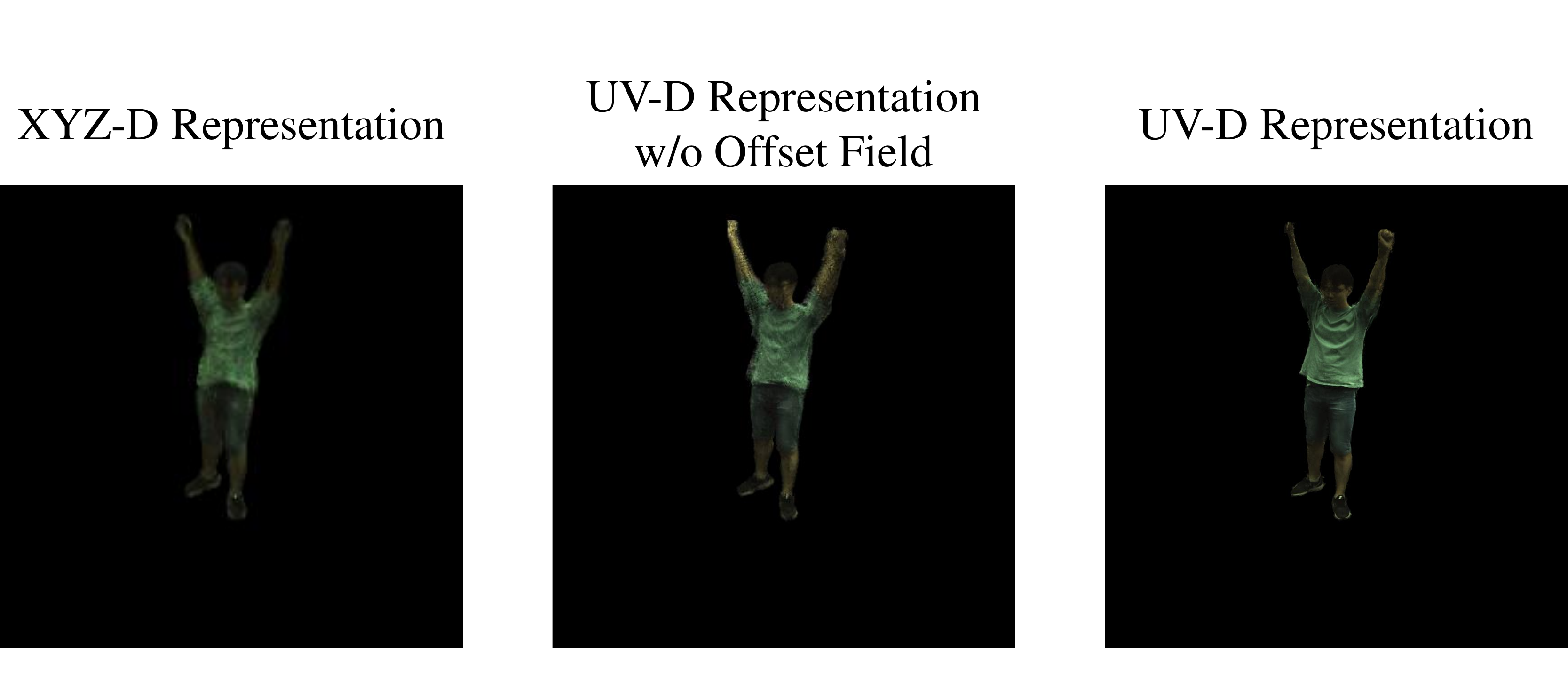}
    \vspace*{-10mm}
    \caption{Ablation study on the UV-D representation and the offset field. ``XYZ-D Representation'' means that the offset field and hash encoding takes the XYZ-D representation as input. ``UV-D Representation w/o Offset Field'' represents that we remove the offset field in our pipeline. ``UV-D Representation'' refers to our full pipeline.}
    \label{fig:ablation_off}

\end{figure}

We attempt to remove the offset field (see Sec\ref{offset}) from our pipeline. 
We performed this ablation on the ZJU-Mocap dataset, choosing the same subjects as for our monocular comparison.
As shown in Fig.~\ref{fig:ablation_off}, without the help of the offset field, our method produces blurry results. And the wrinkles of the clothes turn into average shapes without the offset field. The quantitative result in Tab.~\ref{tab:ablation_off} illustrates that adding an offset field provides further gains. Therefore, we can see that the offset field can help model non-rigid deformation details.

\subsubsection{Ablation Study on the UV-D Representation}
To further evaluate the effectiveness of our UV-D representation, we compare the UV-D representation and the XYZ-D representation defined in Sec.~\ref{UVD representation}.
We also construct an offset field based on the XYZ-D representation $\Delta\mathbf{r'}=F_{\phi'}(\mathbf{r'},\mathbf{e}_{t})$, where $\mathbf{r'}$ is an intrinsic coordinate under the XYZ-D representation.
We performed this ablation on the same data as the ablation on the offset field.
Fig.~\ref{fig:ablation_off} and Tab.~\ref{tab:ablation_off} show the result. Though the XYZ-D representation can aggregate information across frames like the UV-D representation, it is not easy to optimize an offset field based on the XYZ-D representation. Therefore, the XYZ-D represntation cannot handle clothing deformation and produces blurry results. In contrast, our UV-D grid is a smooth and convex domain, we can further refine the results within this domain.
\begin{table}[t]
    \caption{Quantitative result of the ablation study on UV-D representation and offset field. We train these models on ``Camera (1)'' and use the remaining views for testing.}
    \vspace{-4mm}
    \centering
    \begin{tabular}{|c|c|c|c|}
    \hline
          & XYZ-D Representation  & Without offset field  & Full pipeline   \\
    \hline
    PSNR&24.50 &24.42&25.86 \\
    \hline
    SSIM&0.865 &0.863 &0.889 \\
    \hline
    \end{tabular}
    \label{tab:ablation_off}
    \vspace{-4mm}
\end{table}

\begin{figure}
    \centering
    \includegraphics[width=\linewidth]{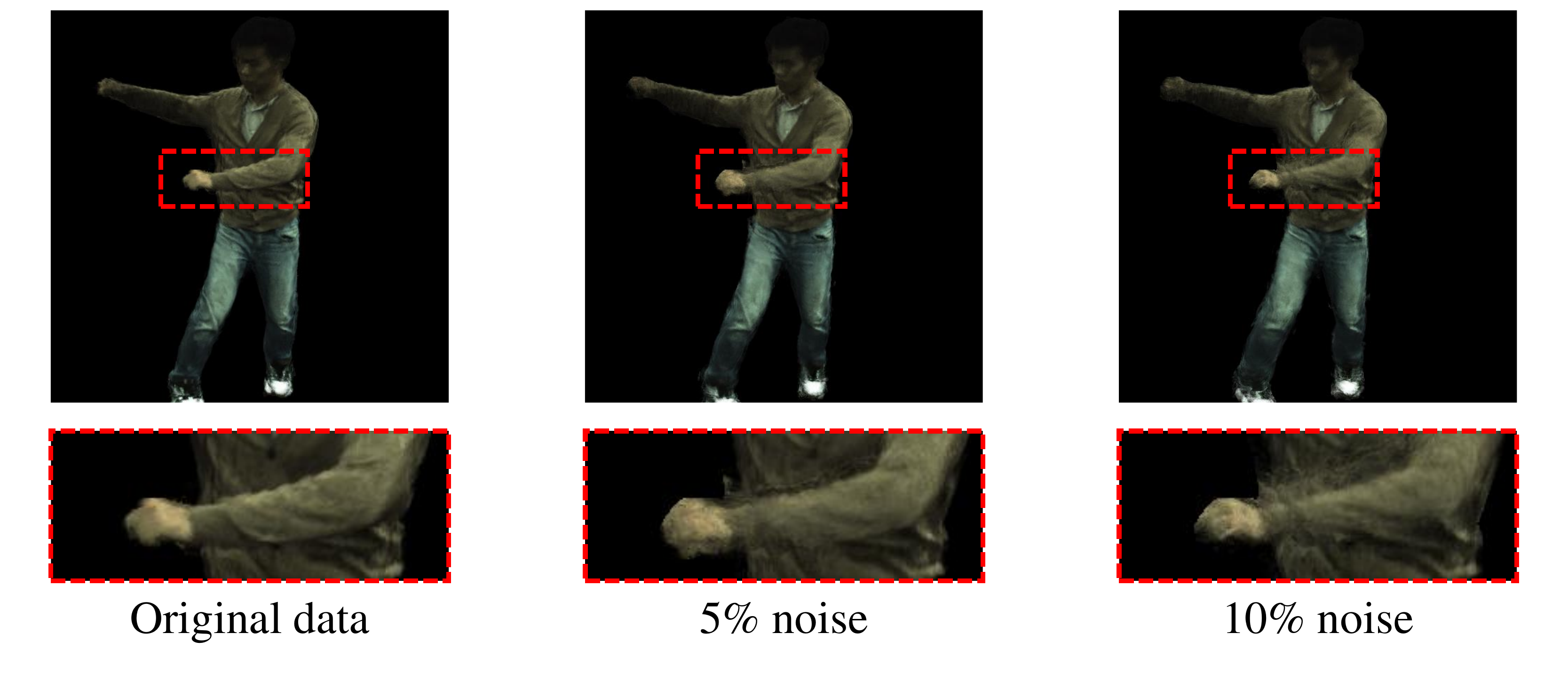}
    \vspace*{-11mm}
    \caption{Ablation study on accuracy of the SMPL parameters. ``5\%'' means that the absolute value of the added noise amount to 5\% of the absolute value of the original parameter.}
    \label{fig:ablation_smpl}
    \vspace*{-4mm}
\end{figure}

\begin{figure*}
    \centering
    \includegraphics[width=\textwidth]{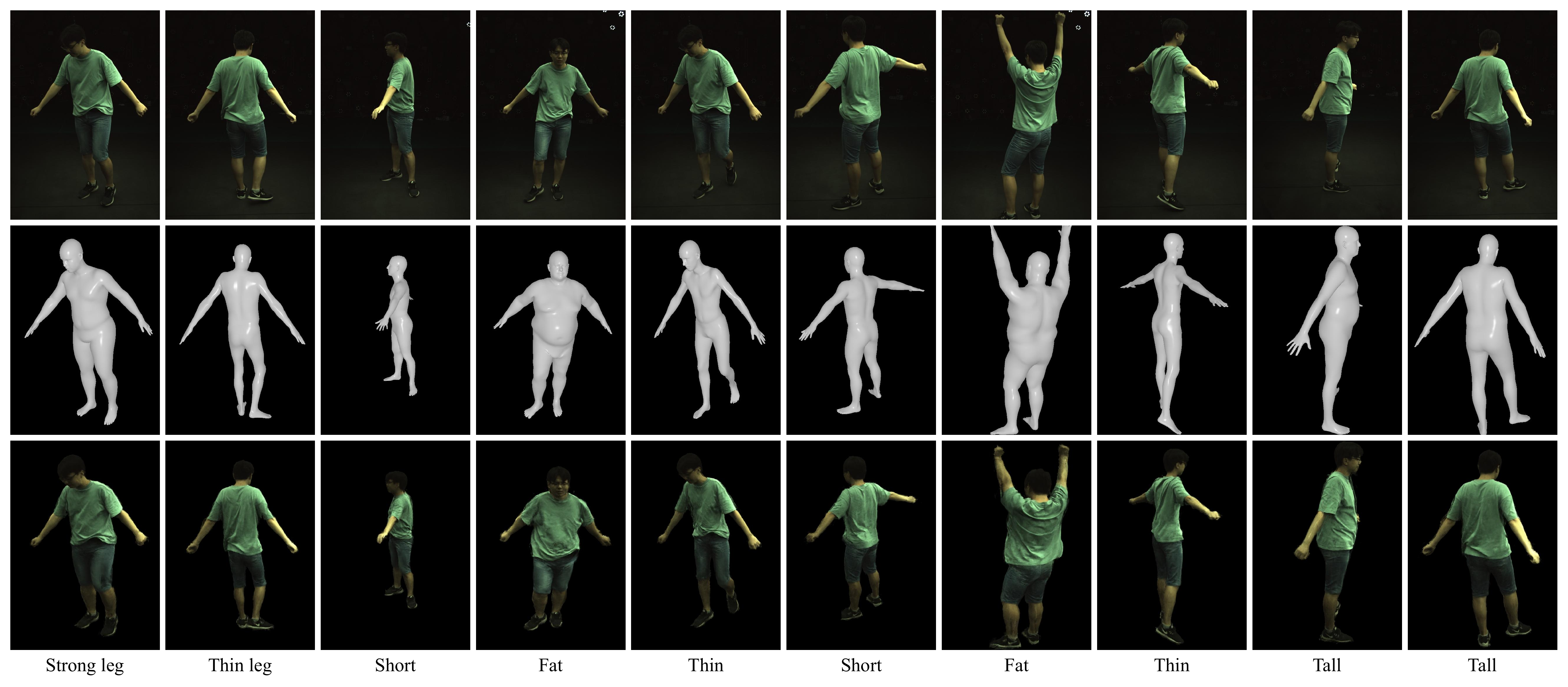}
    \vspace*{-9mm}
    \caption{Application of shape editing.
    Top to bottom: reference video, target shape, and output result.}
    \label{fig:shape_editing}
\end{figure*}

\subsubsection{Ablation Study on the Number of Training Views}
To evaluate the impact of the number of input views on our framework, we compare the results of our method with different input views. 
We did this on the ZJU Mocap dataset, using the same five subjects as in the monocular comparison.
\begin{table}[h]
    \centering
    \caption{Ablation study on the number of training views. We select six camera views and use the remaining views for testing.}
    \begin{tabular}{|c|c|c|c|}
        \hline
             &1 view& 3 views & 6 views \\
        \hline
        PSNR & 25.98 & 27.79 &28.71\\ 
        \hline
        SSIM& 0.893 & 0.905& 0.912\\ 
        \hline
    \end{tabular}
    \vspace{-4mm}
    \label{tab:numcamera}
\end{table}
Tab.~\ref{tab:numcamera} shows the quantitative result, demonstrating that the number of input cameras improves performance on novel views. It should be noted that our method can produce high quality results when trained on monocular inputs.

\subsubsection{Ablation study on Accuracy of the SMPL Parameters}
Although we do not directly optimize the SMPL parameters like HumanNeRF\cite{weng_humannerf_2022_cvpr}, our offset field makes our method less sensitive to the accuracy of the SMPL pose parameters. To demonstrate this, we add random positive and negative noise to the last 69 dimensions of the SMPL pose parameters. We performed this experiment on the ZJU-Mocap dataset on subjects 313, 377, 386, 392, and 394, and compared it to NeuralBody\cite{peng2021neural}, which also does not optimize the SMPL parameters. 
As shown in Fig.~\ref{fig:ablation_smpl}, if the SMPL parameters are inaccurate, our method will produce artifacts in details such as sleeves. 
Tab.~\ref{tab:smpl_noise} shows the quantitative results, and our method is less sensitive to the accuracy of the SMPL parameters than NeuralBody.

\begin{table}[h]
    \begin{center}
    \caption{Ablation study on the accuracy of the SMPL parameters. 5\% means that the absolute value of the added noise amount to 5\% of the absolute value of the original parameter.}
    \vspace{-2mm}
    \resizebox{\linewidth}{!}{
    \begin{tabular}{|c|c|c|c|}
    \hline
          &Original & Add 5\% noise  & Add 10\% noise    \\
    \hline
    NeuralBody&24.15 &23.77 & 23.06      \\
    \hline
    Ours&25.86 &25.59  &25.11          \\
    \hline
    \end{tabular}}
    \label{tab:smpl_noise}
    \end{center}

    \vspace{-4mm}
\end{table}
\subsection{Application: Shape Editing}
Since IntrinsicNGP's representation is based on explicit human surfaces and our UV-D grid can be explained as a 3D density-color map. Therefore, the trained model can be used for shape editing by directly editing the input surfaces. Specifically, we edit the shape parameters of SMPL to generate new geometry proxies and render novel shape results with them. The qualitative results are shown in Fig.~\ref{fig:shape_editing}. Our method produces results that retain the original high-fidelity clothing details and match the input body shape.

\subsection{Visualization of UV-D Grid}
To better demonstrate the meaning of proposed UV-D representation, we made a visualization. We sample rays in the UV-D grid and apply volume rendering to get the rendering results.
\vspace*{-3mm}
\begin{figure}[h]
    \centering
    \includegraphics[width=\linewidth]{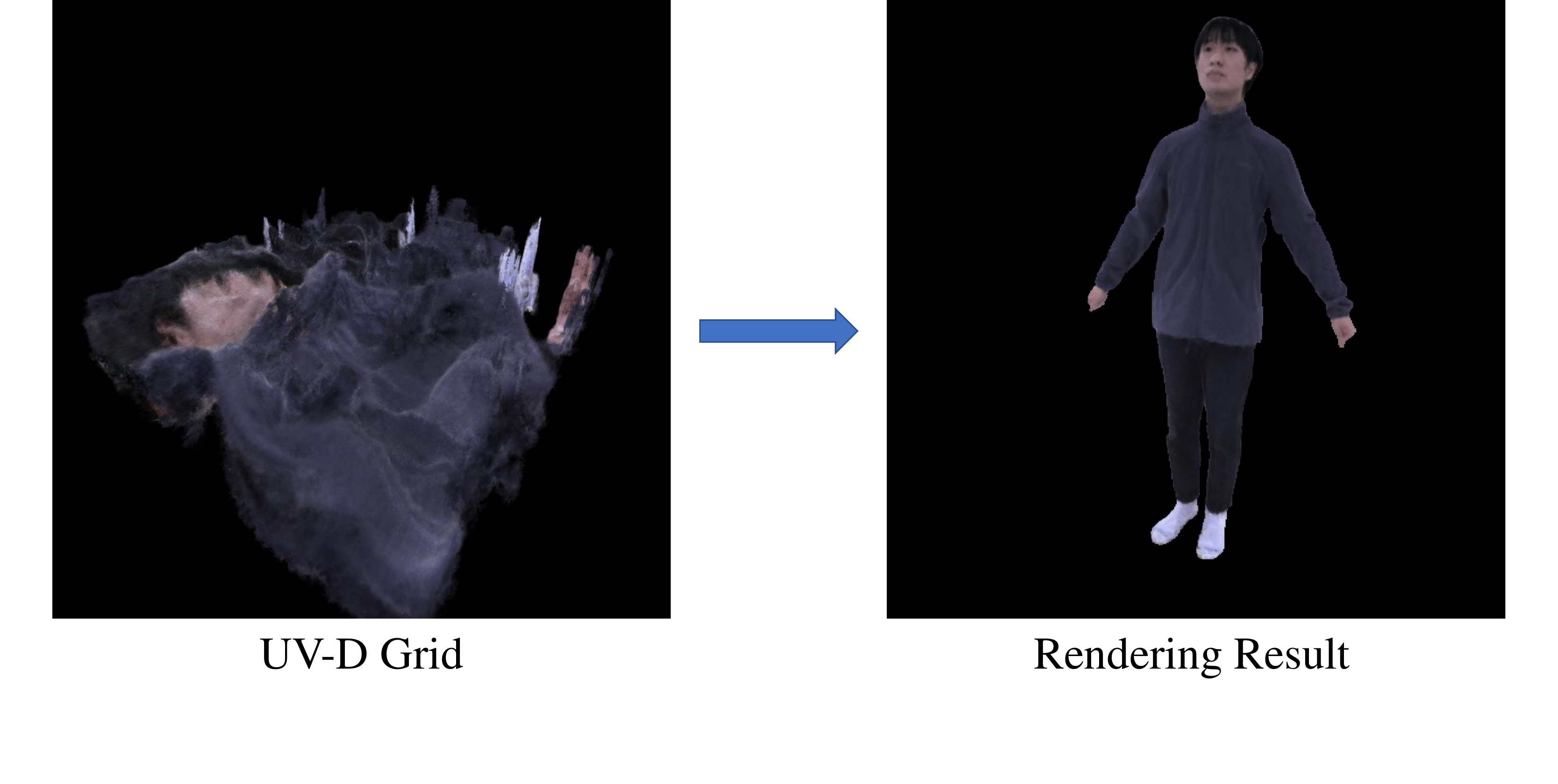}
    \vspace*{-13mm}
    \caption{Visualization of the UV-D grid. 
    }
    \label{fig:visuvd}
    \vspace*{-2mm}
\end{figure}

\begin{table}[h]
    \caption{Analogy of our method to the traditional rendering pipeline.}
    \vspace{-7mm}
    \begin{center}
    \resizebox{\linewidth}{!}{
    \begin{tabular}{|c|c|c|c|c|}
    \hline
          & Geometry proxy & Texture/Color and density  & Query mode &Rendering mode   \\
    \hline
    Traditional&Precise mesh &UV map&UV coordinate&Surface rendering \\
    \hline
    Ours&Rough surface &UV-D grid &Intrinsic coordinate&Volume rendering \\
    \hline
    \end{tabular}}
    \end{center}
    \label{tab:rendering_type}
    \vspace{-3mm}
\end{table}
As shown in Fig.~\ref{fig:visuvd}, the rendering result of the UV-D grid looks like a UV texture map with an additional axis $d$. In this additional axis, our UV-D grid records the distance of the actual surface from the proxy geometry. Just like the UV map records texture information of 2D manifolds, our UV-D grid records color and density information of the space around the proxy geometry. For a given coordinate $(u,v,d)$ in our UV-D grid, it records the information about the point whose nearest point's UV coordinate is $(u,v)$ and whose signed distance value to the human surface is $d$. Thus, our UV-D grid can be thought of as a 3D UV map. For ease of understanding, we have compared our pipeline to the traditional rendering pipeline in Tab.~\ref{tab:rendering_type}. We strongly recommend that readers watch our video for dynamic results.

\section{Limitation and Discussion}
\label{limitation}
\begin{figure}
    \centering
    \includegraphics[width=\linewidth]{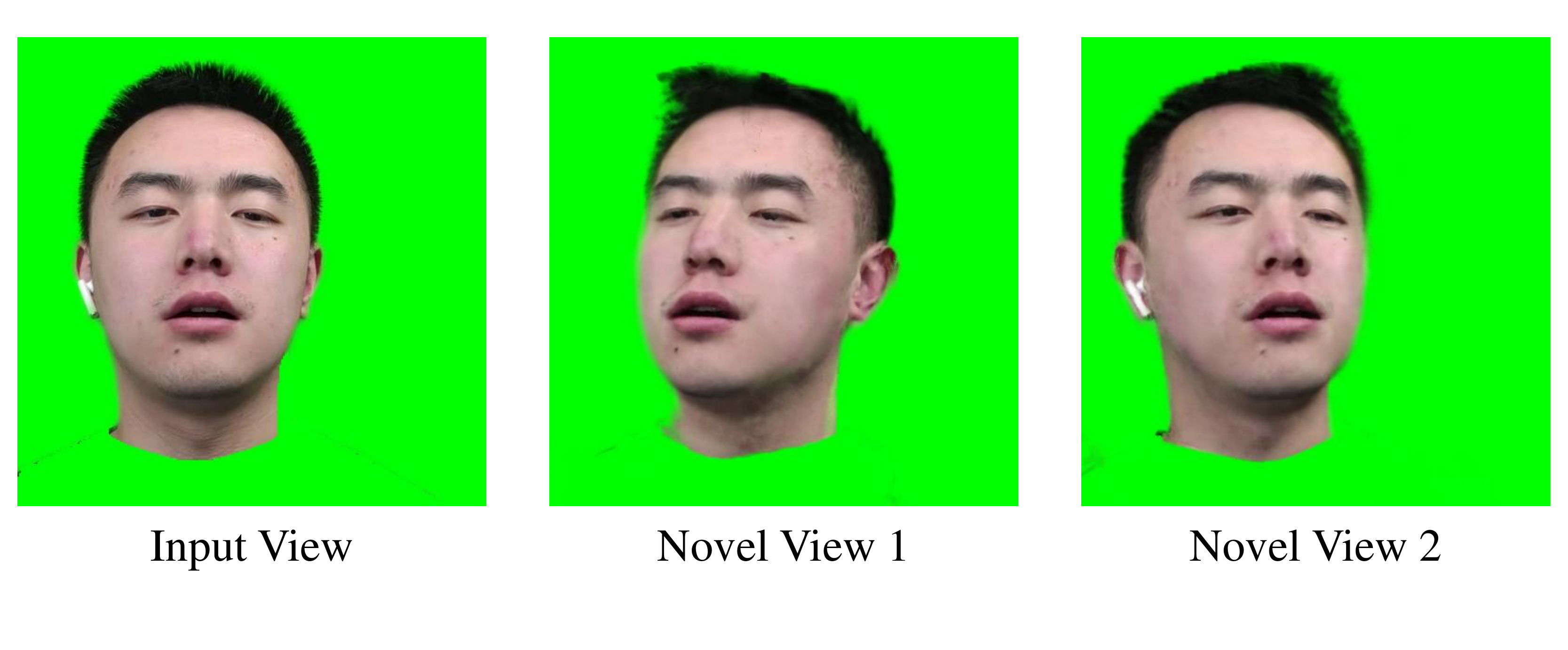}
    \vspace*{-13mm}
    \caption{Novel view synthesis results for the human head.}
    \label{fig:face}
    \vspace*{-4mm}
\end{figure}

\noindent{\bf{Discussion.}} Our proposed intrinsicNGP can not only be used for dynamic human performance, but can also be easily extended to other dynamic objects with corresponding proxy geometry shapes for each frame, such as FLAME\cite{FLAME:SiggraphAsia2017} for human head, SMAL\cite{zuffi20173d} for animals, and non-rigid structure from motion\cite{parashar2021robust} for common objects. To demonstrate this, we made a mini-experiment on human head case. In practice, we capture a monocular video of a talking head and recover the rough surfaces by optimizing the parameters of the FLAME model\cite{FLAME:SiggraphAsia2017}. We then trained on this data for 10 minutes. As shown in Fig.~\ref{fig:face}, our method can achieve high fidelity results and reconstruct details such as the earphone. 

\noindent{\bf{Limitations.}} Currently, our method relies on the template model (in practice, SMPL).
Although we use the offset field to compensate for this, our approach is still limited by the expressiveness of the template model. This problem may be solved with the development of human shape reconstruction methods, as we discussed in Sec\ref{sec:shape}.
\section{Conclusion}
We proposed IntrinsicNGP, an effective and efficient novel view synthesis method for human performers based on the neural radiance field. 
We introduced a novel intrinsic coordinate representation that could aggregate information over time, and extended the hash encoding in INGP from static scenes to dynamic objects. For the captured human performer, IntrinsicNGP could converge to high-quality results within a few minutes. 
Extensive experimental results have shown that we can produce high-quality results for this challenging task, demonstrating potential practical applications of IntrinsicNGP.

\noindent{\bf{Potential Negative Impact}}
\label{societal}
The misuse of our method may raise ethical issues by reconstructing and generating a shape-edited human body without permission. Therefore, we require that the media generated by our method be clearly labeled as synthetic. 
\ifCLASSOPTIONcompsoc
  \section*{Acknowledgments}
\else
  \section*{Acknowledgment}
\fi
This work was supported by the National Natural Science Foundation of China (No. 62122071, No. 62272433).


\newpage

{
\bibliographystyle{IEEEtran}
\bibliography{egbib}
}

\begin{IEEEbiography}[{\includegraphics[width=1in,height=1.25in,clip,keepaspectratio]{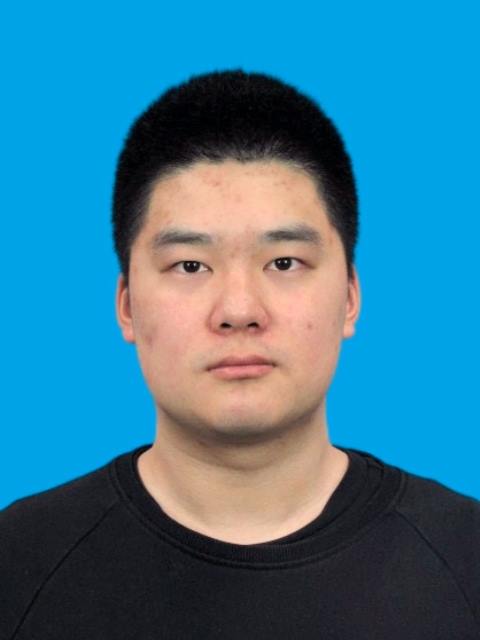}}]{Bo Peng}
is a master student at the
School of Mathematical Sciences,
University of Science and Technology of China. His research interests include human reconstruction and neural rendering.
\end{IEEEbiography}

\begin{IEEEbiography}[{\includegraphics[width=1in,height=1.25in,clip,keepaspectratio]{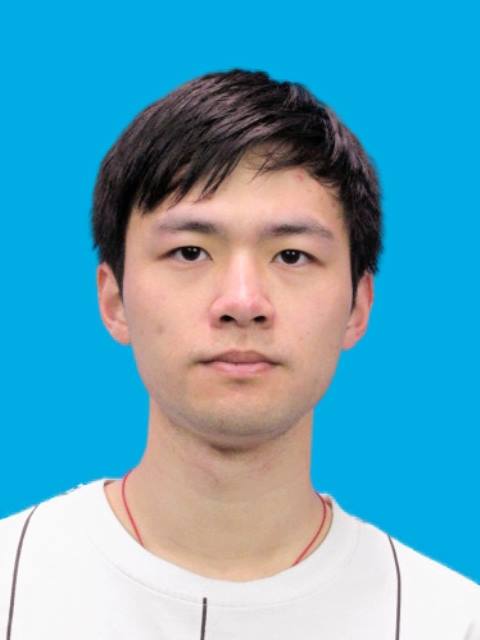}}]{Jun Hu}
is a master student at the School of Mathematica Sciences, University of Science and Technology of China. His research interests include multi-view human synthesis in computer vision.
\end{IEEEbiography}

\begin{IEEEbiography}[{\includegraphics[width=1in,height=1.25in,clip,keepaspectratio]{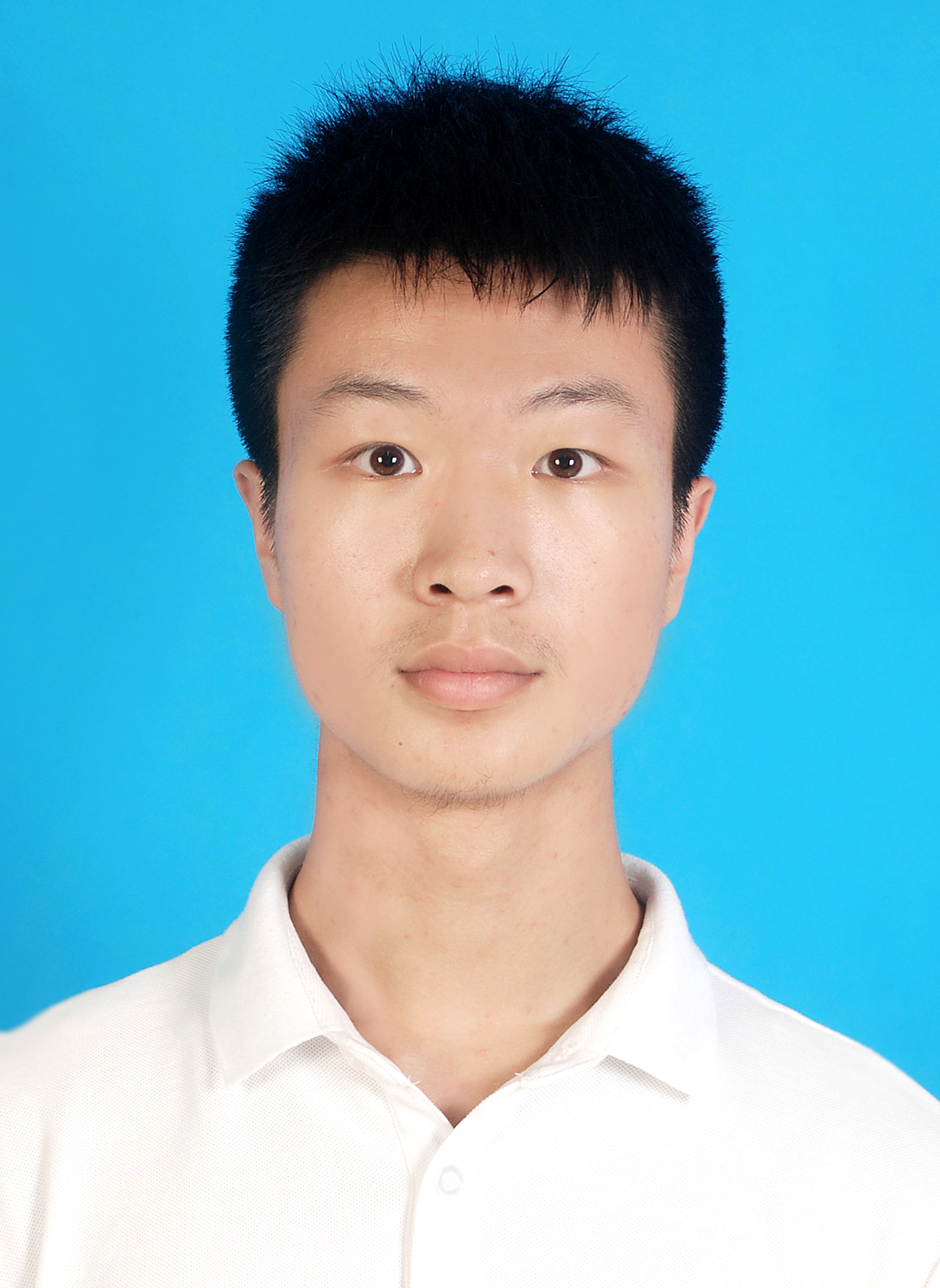}}]{Jingtao Zhou}
is a master student at the School of Mathematical Sciences, University of Science and Technology of China. His research interests include computer vision and deep learning.
\end{IEEEbiography}

\begin{IEEEbiography}[{\includegraphics[width=1in,height=1.25in,clip,keepaspectratio]{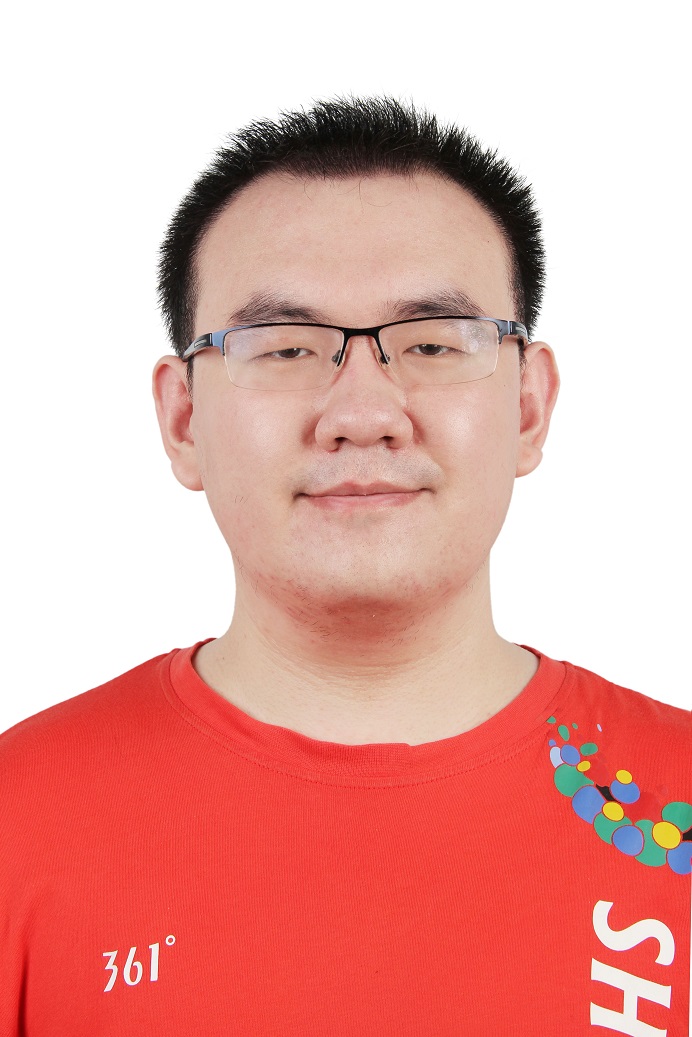}}]{Xuan Gao}
is a Ph.D. student at the School of Mathematical Sciences, University of Science and Technology of China. His research interests include computer graphics and computer vision..
\end{IEEEbiography}

\begin{IEEEbiography}[{\includegraphics[width=1in,height=1.25in,clip,keepaspectratio]{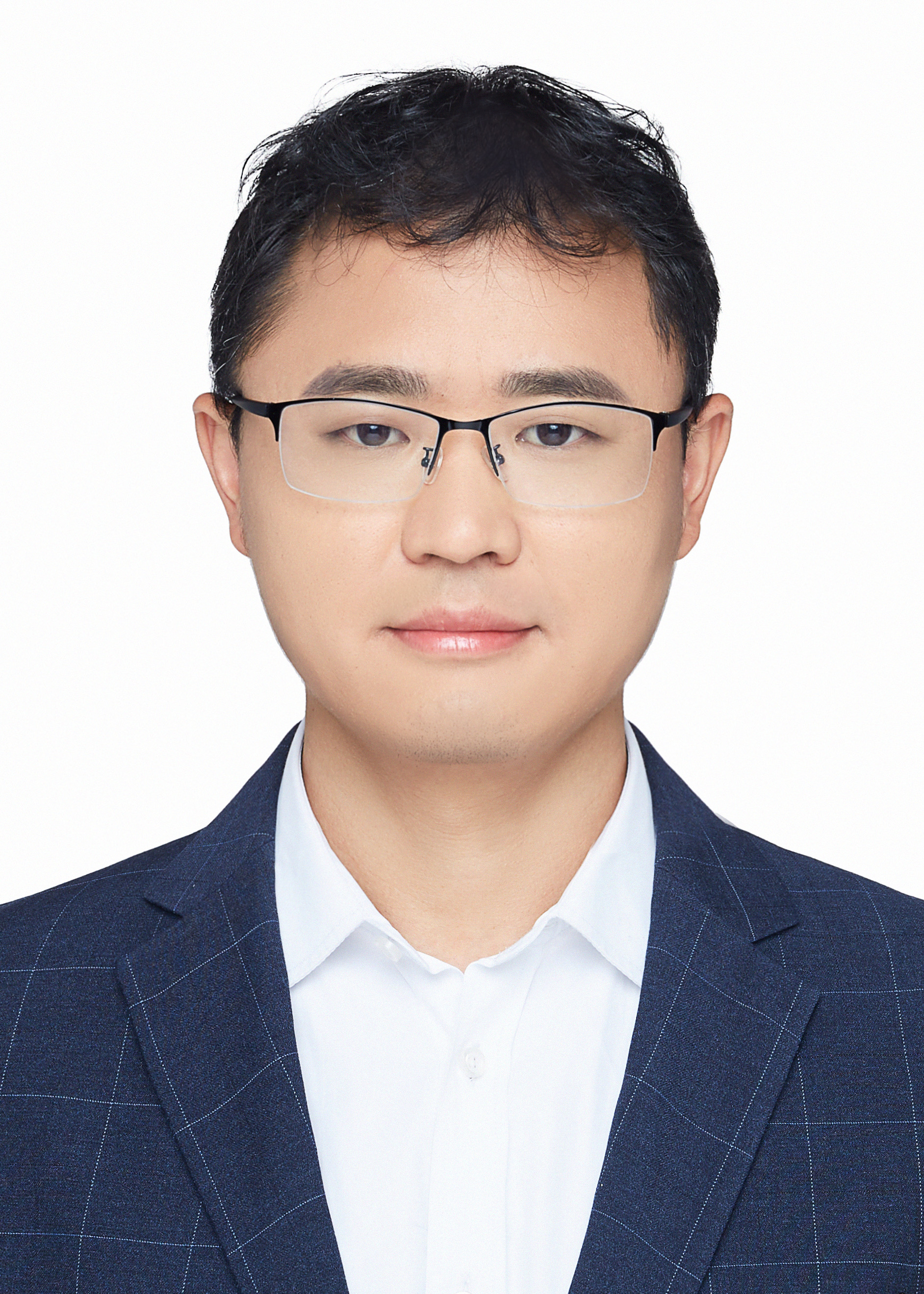}}]{Juyong Zhang}
is a professor in the School of Mathematical Sciences at University of Science and Technology of China. He received the BS degree from the University of Science and Technology of China in 2006, and the Ph.D degree from Nanyang Technological University, Singapore. His research interests include computer graphics, computer vision, and numerical optimization. He is an associate editor of IEEE Transactions on Multimedia.
\end{IEEEbiography}

\end{document}